\documentclass[journal,compsoc]{IEEEtran}
\ifCLASSOPTIONcompsoc
  \usepackage[nocompress]{cite}
\else
  \usepackage{cite}
\fi

\ifCLASSINFOpdf
 \else
 \fi
 
\usepackage{amsmath,amsfonts}
\usepackage{array}
\usepackage[caption=false,font=normalsize,labelfont=sf,textfont=sf]{subfig}
\usepackage{textcomp}
\usepackage{stfloats}
\usepackage{url}
\usepackage{colortbl}
\usepackage{verbatim}
\usepackage{graphicx}
\usepackage{cite}
\usepackage[utf8]{inputenc}
\usepackage[T1]{fontenc}
\usepackage{hyperref}
\usepackage{url}
\usepackage{booktabs}
\usepackage{amsfonts}
\usepackage{nicefrac}
\usepackage{microtype}
\usepackage{xcolor}
\usepackage{threeparttable}
\usepackage{color}
\usepackage{makecell}
\usepackage{multirow}
\usepackage{graphicx}
\usepackage{wrapfig}
\usepackage{algorithmic}
\usepackage{algorithm}
\usepackage{enumitem}
\usepackage{ragged2e}

\newcolumntype{L}[1]{>{\raggedright\let\newline\\\arraybackslash\hspace{0pt}}m{#1}}
\newcolumntype{C}[1]{>{\centering\let\newline\\\arraybackslash\hspace{0pt}}m{#1}}
\newcolumntype{R}[1]{>{\raggedleft\let\newline\\\arraybackslash\hspace{0pt}}m{#1}}

\begin{document}

\title{Deep Temporal Graph Clustering:\\A Comprehensive Benchmark and Datasets}

\author{Meng Liu,
Ke Liang,
Siwei Wang, Xingchen Hu,\\ Sihang Zhou, Xinwang Liu,~\IEEEmembership{Senior~Member,~IEEE}
\IEEEcompsocitemizethanks{
\IEEEcompsocthanksitem Corresponding Authors: Xinwang Liu, Siwei Wang.
\IEEEcompsocthanksitem This work was supported by the National Natural Science Foundation of China (no. 62325604, 62276271, 62441618, 62376279). 
\IEEEcompsocthanksitem Meng Liu, Ke Liang, and Xinwang Liu are with the College of Computer Science and Technology, National University of Defense Technology, Changsha, 410073, China. E-mail: {mengliuedu@163.com, xinwangliu@nudt.edu.cn}.
\IEEEcompsocthanksitem Siwei Wang is with the Intelligent Game and Decision Lab, Beijing, 100091, China. E-mail: {wangsiwei13@nudt.edu.cn}.
\IEEEcompsocthanksitem Xingchen Hu is with the College of Systems Engineering, National University of Defense Technology, Changsha, 410073, China.
\IEEEcompsocthanksitem Sihang Zhou is with the College of Intelligence Science and Technology, National University of Defense Technology, Changsha, 410073, China.
}}

\IEEEtitleabstractindextext{%
\begin{abstract}
\justifying
Temporal Graph Clustering (TGC) is a new task with little attention, focusing on node clustering in temporal graphs. Compared with existing static graph clustering, it can find the balance between time requirement and space requirement (Time-Space Balance) through the interaction sequence-based batch-processing pattern. However, there are two major challenges that hinder the development of TGC, i.e., inapplicable clustering techniques and inapplicable datasets. To address these challenges, we propose a comprehensive benchmark, called BenchTGC. Specially, we design a BenchTGC Framework to illustrate the paradigm of temporal graph clustering and improve existing clustering techniques to fit temporal graphs. In addition, we also discuss problems with public temporal graph datasets and develop multiple datasets suitable for TGC task, called BenchTGC Datasets. According to extensive experiments, we not only verify the advantages of BenchTGC, but also demonstrate the necessity and importance of TGC task. We wish to point out that the dynamically changing and complex scenarios in real world are the foundation of temporal graph clustering. The code and data is available at: \url{https://github.com/MGitHubL/BenchTGC}.
\end{abstract}

}

\maketitle
\IEEEdisplaynontitleabstractindextext
\IEEEpeerreviewmaketitle

\ifCLASSOPTIONcompsoc
\IEEEraisesectionheading{\section{Introduction}\label{sec:introduction}}
\else
\section{Introduction}
\label{sec:introduction}
\fi

\IEEEPARstart{G}{raph} clustering is one of the key branches of clustering, which aims to dividing nodes in the graph into different clusters \cite{pan2021multi, zhang2023large, mo2023multiplex}. Since real-world data often lacks label information, as a classic task in unsupervised scenarios, clustering can be used in many real-world applications, such as community discovery and anomaly detection \cite{jin2022towards, yang2022robust}. It can also assist with recommendation systems and bioinformatics analysis \cite{sheng2023multi}.

The emergence of deep learning has made deep graph clustering begin to attract more attention. Existing methods usually learn a representation (also known as embedding) for each node, and these node embeddings are used to perform clustering tasks \cite{li2021adaptive, wang2020large}. Note that existing methods almost only focus on node clustering on static graphs, while ignoring the dynamic evolution and temporal information in the graph \cite{wang2023network, wu2024high}. In some scenarios, this can have unpredictable negative consequences.

\begin{figure}[t]
    \centering
\includegraphics[width=0.48\textwidth]{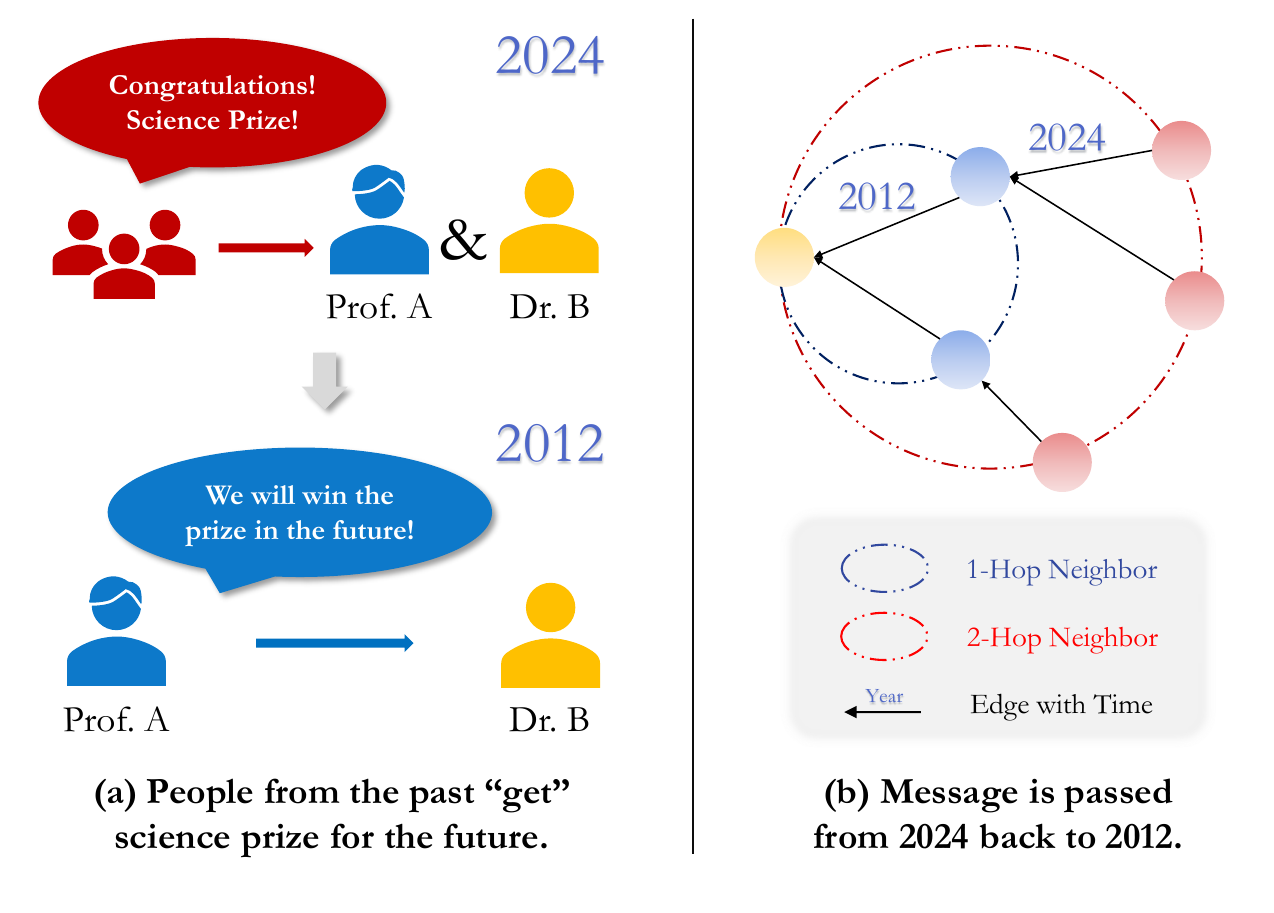}
    \caption{The knowledge leakage problem in the message passing mechanism of graph learning. Figure (a): The professor knew as early as 2012 that he would win the award in 2024. Figure (b): In the message passing mechanism, information will be propagated backwards from the future to the past.}
    \label{intro}
    \vspace{-3mm}
\end{figure}

Imagine such a scene in Figure \ref{intro}(a), professor A and his student Dr. B jointly won an academic award in 2024. After learning the news, Prof. A told Dr. B as early as 2012: ``\emph{Work hard and we will definitely win the award in the future!}'' This scene may be seen in science movies, but it is impossible to happen in the real world. Because information propagation in the real world follows a basic rule, that is, \textbf{the past cannot predict the future}. However, in the classic message passing mechanism of deep graph clustering, this rule is violated. 

In Figure \ref{intro}(b), we abstract the above scene into a graph structure. During the message passing process in the graph, neighborhood information will be aggregated to the source node, and this process will go through multi-hop neighborhoods. In this case, a message will be passed by the red node to the blue node through the edge established in 2024, and then passed to the yellow node through the edge established in 2012. It means that knowledge leaks undoubtedly occurred. In other words, the classic message passing mechanism of graph learning may cause reverse information flow, resulting in unpredictable errors.

Furthermore, considering the graph clustering task, if we want to consider what field a professor is currently focusing on, do we need to consider the papers he published in the past 3 years or the articles he published 20 years ago? If the static graph clustering methods are used without considering time information, it is very likely that the professor will be assigned to the wrong field because these methods give equal importance to the above papers. In fact, for clustering task, when the latest results are needed, we should pay more attention to the papers in the past 3 years, and when comprehensive results are needed, we should also consider the papers 20 years ago.

In this case, time information is particularly important, thus temporal graph learning came into being. As an important branch of graph learning that focuses on time information, temporal graph learning no longer use adjacency matrix but instead use interaction sequence to store graph data. Such sequence stores node interactions in time and can be divided into multiple batches for training. Due to its focus on real-world dynamic environments, temporal graph learning has gradually attracted attention.

However, current temporal graph works always focus on link prediction task, but ignore another classic task in unsupervised scenarios, i.e., node clustering. \textbf{Thus we ask: What makes temporal graph clustering (TGC) ignored?}

We argue that there are two major challenges that hinder the development of temporal graph clustering. \textbf{(1) Inapplicable clustering techniques.} Due to the interaction sequence-based data structure, temporal graph learning usually utilize batch-processing pattern. In a single batch, there are no adjacency matrix can be used and matched, which further causes many adjacency matrix-based clustering techniques to no longer be applicable. This brings difficulties to researchers, i.e., they need to design new clustering modules for temporal graph scenarios. \textbf{(2) Inapplicable datasets.} Many existing public temporal graph datasets lack labels. Although the node clustering task does not require labels in training, labels still need to be used to evaluate performance. This also brings difficulties to researchers, i.e., even if there is a temporal graph clustering method, it is difficult to find some suitable datasets for evaluation.

To solve these two challenges, we propose BenchTGC, a comprehensive benchmark for temporal graph clustering. In particular, we design a 3-stage framework and develop a set of datasets for TGC task. The \textbf{BenchTGC Framework} consists of 3 stages, i.e., pre-processing, training, and clustering. We discuss in detail the different technical means involved in these stages, and focus on improving multiple classic graph clustering techniques in training stage for use in temporal graph methods. Such framework can be flexibly deployed to almost all temporal graph methods and bring about a significant improvement in node clustering performance. The \textbf{BenchTGC Datasets} contains 3 public temporal graph datasets and 6 large-scale datasets that we further developed. We also focus on why most existing public temporal graph datasets are not suitable for TGC task, and divide these datasets into three cases: label-unavailable, clustering-inapplicable , and clustering-applicable. We refine and extract real-world data to form 6 new datasets for TGC task, effectively filling the gap in the community.

Finally, we compare BenchTGC framework with 16 existing methods on BenchTGC datasets and conduct extensive experiments to illustrate the difference between temporal methods and static methods on node clustering, as well as the potential benefits that TGC can bring. We point out that temporal graph clustering is an effective supplement to static graph clustering, and it provides researchers with a new solution because TGC can maintain the balance between time and space requirements, i.e., \textbf{``Time-Space Balance''}.

This paper is an extended version of the conference paper which published in ICLR 2024 main track \cite{TGC_ML_ICLR}. To the best of our knowledge, we are the first work which comprehensively discuss deep temporal graph clustering. Such conference version won the Best Paper Award of 2024 China Computational Power Conference. Compared with our conference version, our journal version has been substantially expanded by more than 50\%, its contributions come from multiple perspectives:

\textbf{New Framework.} Compared to the conference version, we propose a more comprehensive 3-stage framework that can be deployed on almost all temporal graph methods. In the framework, we improve more clustering techniques to accommodate the interaction sequence-based batch-processing pattern, and give the definition and paradigm for temporal graph clustering.

\textbf{New Datasets.}
We systematically analysis why most existing datasets are not suitable for TGC task, and divide them into: unlabeled, clustering unavailable, and clustering available. Furthermore, we propose Data4TGC which includes 3 public datasets and 6 developed large-scale datasets. These developed datasets have larger scale and accurate node labels than most public datasets.

\textbf{More Experiments and Better Performance.} We conduct more experiments and significantly improve the performance. Specially, we achieve the best 42.15\% improvement on ``arXivCS'' dataset  (ARI metric from 24.65\% to 35.04\%), and 100\% performance of 4 metrics on School dataset. Combined with experiments, we point out that TGC can be an effective supplement to static graph clustering because it can find the balance between time requirement and space requirement, i.e., ``Time-Space Balance''.

\textbf{More Challenges Discussion.}
We point out the challenges that hinder the development of TGC, i.e., inapplicable clustering techniques and inapplicable datasets. We also present the limitations and applications of TGC, and argue that the dynamically changing and complex scenarios in real world are the foundation of TGC.

The rest of this paper is organized as follows. In Section 2, we introduce the related works. In Section 3, we give the details of our BenchTGC Framework. In Section 4, we discuss the public datasets and developed BenchTGC Datasets. In Section 5, we report the experimental results and analyses. In Section 6, we discuss the limitations and applications. In Section 7, we conclude our work.

\section{Related Work}

In this part, we first introduce existing deep graph clustering methods, then present temporal graph learning methods. We also discuss some works that similar but different from temporal graph clustering. 

\subsection{Deep Graph Clustering}

Graph clustering, also referred to node clustering, represents a fundamental unsupervised task within the realm of graph learning. Graph clustering can be utilized in many real-world applications. In recent times, there has been a notable emergence of deep learning methodologies employed for the purpose of graph clustering. This trend has led to the proliferation of a plethora of methodologies and techniques in this field of research. Some works perform clustering tasks on graph data, and some use graph techniques to solve clustering problems \cite{ren2019structured, luo2019discrete}.

For instance, DNGR \cite{cao2016deep} extracts a low-dimensional embedding for each node, effectively capturing the structural information. DAEGC \cite{wang2019attributed} encodes the structure and features into a compact representation using an attention network. ARGA \cite{pan2018adversarially} encodes both the graph structure and node information into a concise embedding. MVGRL \cite{hassani2020contrastive} is a self-supervised method that generates node representations by contrasting structural views of graphs. AGE \cite{cui2020adaptive} applies a Laplacian smoothing filter, carefully designed to enhance the filtered features to obtain improved node embeddings. SDCN \cite{bo2020structural} effectively integrates structural information into the process of deep clustering. DFCN \cite{tu2021deep} utilizes two sub-networks to independently process augmented graphs. DCRN \cite{liu2022DCRN} proposes a dual correlation reduction module to decrease information correlation in a dual manner. CGC \cite{park2022cgc} employs a contrastive framework to simultaneously learn node embeddings and cluster assignments. SCGC \cite{liu2023simple} designs a straightforward data argumentation network for rapid node clustering.

The majority of these methods are based on static graphs. However, information in the real world is often dynamic. This leads to the concept of temporal graph, which emphasizes time information in nodes interactions.

\subsection{Temporal Graph Learning}

Graph data can be classified into static and dynamic graphs based on the absence or presence of time information. Traditional static graphs represent data in the form of an adjacency matrix, where samples are nodes and relationships between samples are edges \cite{he2021adversarial}. Dynamic graphs can be further classified into discrete graphs (discrete-time dynamic graphs, DTDGs) and temporal graphs (continuous-time dynamic graphs, CTDGs). Discrete graphs are composed of multiple static snapshots that are based on fixed time intervals. Each snapshot is regarded as a distinct static graph, and these snapshots are arranged in chronological order \cite{gao2021equivalence, liang2022survey}. Discrete graph methods typically use the static model to learn each snapshot and then introduce RNN or attention modules to capture the time information between different snapshots, such as EvolveGCN \cite{pareja2020evolvegcn} and DySAT \cite{sankar2020dysat}. In this scenario, conventional graph clustering techniques can still be applied to handle discrete graphs effectively\footnote{Since these methods and static graph methods are all constructed based on the adjacency matrix, we regard them as the same category as static graph methods in this paper.}.

Unlike discrete graphs, \textbf{temporal graphs} can observe each node interaction more clearly \cite{liu2025rethinking}. Temporal graphs discard the adjacency matrix form and record node interactions based on the sequence directly. Based on the different ways of capturing temporal information, temporal graph methods can be divided into multiple categories. Some methods utilize random walk to model the temporal graph evolution, such as CTDNE \cite{nguyen2018continuous}, and CAW \cite{wang2021inductive}. Some methods consider graph neural network, including TGAT \cite{xu2020inductive}, and TREND \cite{wen2022trend}. Some methods introduce the Hawkes process to capture the historical influence, such as HTNE \cite{zuo2018embedding} and MNCI \cite{MNCI_ML_SIGIR}. Some methods are devoted to exploring the evolution of node representations, such as DyRep \cite{trivedi2019dyrep} and TGN \cite{rossi2020temporal}. In addition, several methods dedicate to exploring new boundaries. For example, JODIE \cite{kumar2019predicting} attempts to directly predict node representations at future moments. OTGNet \cite{feng2023towards} considers temporal graph learning in the open-world.

The temporal graph methods introduced above almost all focus on link prediction, while ignoring node clustering. There are also some works related to the concept of temporal are worth learning from, such as \cite{marcu2020semantics} and \cite{marcu2023self} treat the propagation of segmentation labels through time as a spacetime clustering problem. However, the above works are essentially different from temporal graph clustering.

\subsection{Similar Works to TGC}

As mentioned above, deep temporal graph clustering, as a new task definition, has received little attention from researchers. Although there are some works that seem similar to deep temporal graph clustering, they are actually significantly different from what we are talking about:

(1) Although CGC \cite{park2022cgc} purports to perform experiments on temporal graph clustering, it actually conducts experiments solely on discrete dynamic graphs using only one dataset. Although discrete graphs and temporal graphs can be converted into each other, discrete graphs need to be treated as static graphs for each snapshot, which makes it challenging to handle large-scale graphs efficiently. In other words, even discrete dynamic graphs (equivalent to static graphs) with only one static snapshot can cause memory overflow problems. As the same reason, DNE \cite{du2018dynamic}, RTSC \cite{you2021robust}, DyGCN \cite{cui2021dygcn}, and VGRGMM \cite{li2022exploring} are successful on discrete graphs, but not applicable to temporal graphs.

(2) Despite the title of GRACE \cite{yang2017graph} including ``dynamic embedding'', it primarily pertains to dynamic self-adjustment, which is still based on static graph analysis. STAR \cite{xu2019spatio} and Yao et al. \cite{yao2021interpretable} concentrate on node classification, which does not align with node clustering.

(3) Some methods claim to discuss clustering on dynamic or temporal graphs \cite{gorke2009dynamic, gorke2013dynamic, matias2017statistical, ruan2021dynamic}, but they use traditional techniques instead of deep learning techniques, that we do not discuss here.

(4) Some works contain keywords such as ``temporal / dynamic graph clustering'' in the title but are less relevant to temporal graph data, but are more about applying the concept of ``dynamic'', such as DeGTeC \cite{liang2023degtec}, DCFG \cite{bu2017dynamic}, and DPC-DLP \cite{seyedi2019dynamic}, etc.

Through these related works, we argue that the TGC task receives little attention. With this motivation, we propose the BenchTGC benchmark to further introduce it.

\section{BenchTGC Framework}

\begin{figure*}[t]
    \centering
\includegraphics[width=0.98\textwidth]{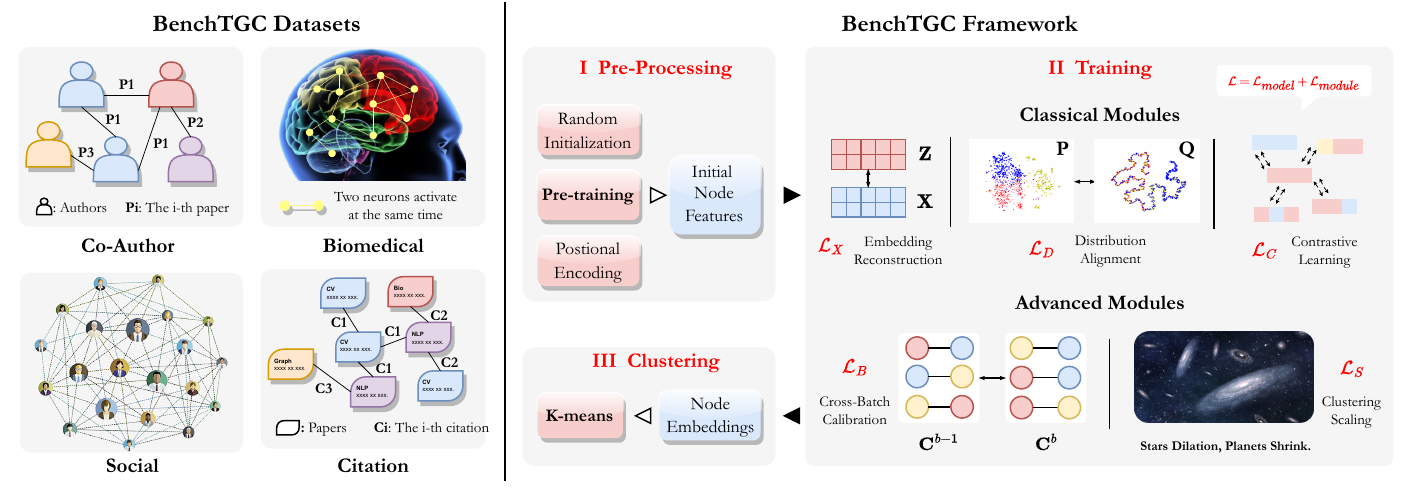}
    \caption{Overview of BenchTGC. BenchTGC includes a set of datasets and 3-stage framework. Such datasets are developed from real-world scenes and covers multiple areas. Such framework divides the temporal graph clustering task into 3 stages: pre-processing, training, and clustering.}
    \label{framework}
\end{figure*}

In this section, we propose the \textbf{BenchTGC Framework} to address the first challenge affecting the development of TGC, i.e., \textbf{Inapplicable clustering techniques}.

\subsection{Overview}

As shown in Figure \ref{framework}, we introduce the real-world datasets used in TGC task and 3-stage procedure of BenchTGC Framework. The sources of our BenchTGC Datasets are diverse and distributed in various real-world scenarios, which will be introduced in next section. For the BenchTGC Framework, we divide it into 3 stage: pre-processing, training, and clustering. There are different coping strategies at different stages. In addition, we also discuss complexity comparison of graph clustering.

\subsection{Problem Definition}

Here we give the definitions of temporal graph and temporal graph clustering.

\textbf{\emph{Definition 1.} Temporal graph.} If a graph records the node interactions at each timestamps as a sequence, it is called a temporal graph. Given a temporal graph $\mathcal{G}=(\mathcal{V}, \mathcal{E}, \mathcal{T})$, $\mathcal{V}$ is the set of nodes, $\mathcal{E}$ is the set of node interactions, and $\mathcal{T}$ records the timestamp of each interaction. The concept of edge in static graph is replaced by interaction in temporal graph, because multiple interactions may occur at different timestamps on an edge between two nodes. A temporal graph is stored as an interaction sequence, which includes many interactions \emph{(node, node, time)} ordered by time.

Researchers usually focus on link prediction in temporal graphs, but rarely on temporal graph clustering.

\textbf{\emph{Definition 2.} Temporal graph clustering.} When node clustering is performed on temporal graphs, it is called temporal graph clustering. Node clustering in the graph follows some rules: (1) nodes in a cluster are densely connected, and (2) nodes in different clusters are sparsely connected. Here we define $K$ clusters to divide all nodes, i.e., $\mathbf{C} = \{\mathbf{c}_1, \mathbf{c}_2, ..., \mathbf{c}_k\}$. Node embeddings are continuously optimized during training and then fed into the K-means algorithm for performance evaluation.

\subsection{Paradigm of Temporal Graph Learning}

Before discuss the 3-stage procedure, we first present the paradigm of temporal graph methods. Because our BenchTGC Framework can be deployed on almost any temporal graph method, we do not discuss the design for a particular temporal method.

Given a temporal graph $\mathcal{G}=(\mathcal{V}, \mathcal{E}, \mathcal{T})$, the goal of temporal graph learning is to learn better node embeddings $\mathbf{Z}$ from the graph through the mapping function $\mathcal{F}$ as
\begin{equation} 
    \mathbf{Z} = \mathcal{F}(\mathcal{G}).
\end{equation}
Researchers mainly focus on the design of $\mathcal{F}$, some by introducing the Hawkes process \cite{hawkes1971point}, some by designing dynamic GNNs, some by using RNNs, and so on. The loss function is essential in the design of mapping function $\mathcal{F}$. For temporal graph methods, which often focus on unsupervised link prediction task, their loss functions tend to use positive and negative sample comparisons, i.e.,
\begin{equation} 
    \mathcal{L}_{\operatorname{\emph{model}}} = - \log (\sigma(-\mathbf{z}_i^{\mathsf{T}} \mathbf{z}_j)) - \log(\sigma(\mathbf{z}_i^{\mathsf{T}} \mathbf{z}_n)),
\end{equation}
where node $i$ and $j$ are considered as positive sample pair that is brought closer, and $i$ and $n$ are considered as negative sample pair that is pushed farther apart. Certainly, there are many other loss functions, but we not go into them here, and uniformly denote them as $\mathcal{L}_{\operatorname{\emph{model}}}$.

In actual training, the temporal graph is divided into multiple batches, and model is optimized for each batch, i.e., at the same time, only one single batch $\mathcal{B}$ is used for training, not all of the interactions $\mathcal{E}$. This is the most important pattern of temporal graph learning, and consequently, our improvements to clustering techniques are designed in terms of one single batch.

\subsection{Pre-Processing}

In many public temporal graph datasets, there are no initial features for training. But almost all methods need some information as input (even if just a placeholder), so we need to generate features for the graph before training.

\subsubsection{Feature Generation}

(1) \textbf{Random initialization.} Some methods utilize random initialization to generate features that are intended for placeholder use only and contain no specific meaning. During training, such features (or transformation function to these features) are treated as parameters to be optimized together.

(2) \textbf{One-hot embedding.} In order to express the positional information between samples, researchers perform one-hot embedding based on the sample id, i.e., the $\mathcal{N} \times \mathcal{N}$ vectors are constructed based on the number of samples, and each sample corresponded to a $1 \times \mathcal{N}$ vector. In this vector, the value is 1 only at the position corresponding to that sample id, and 0 for all others. Thus the $i$-th one-hot embedding $\mathbf{z}^0_i$ can be formulated as
\begin{equation} 
    \mathbf{z}^0_i = \{0,\cdots 0,i,0,\cdots,0\}.
\end{equation}

(3) \textbf{Positional encoding.} One-hot embedding, while capable of expressing location information, is too lengthy when faced with a large number of samples. Based on this, researchers propose the positional encoding\cite{vaswani2017attention}, which generates its corresponding $d$-dimensional vector for each sample ID by introducing Fourier transform mechanism. There are different transformation ways for the odd ($2m$) and even ($2m+1$) dimensions of a node $i$, i.e.,
\begin{equation}
  \begin{aligned}
   \textbf{P}(i,2m) & = \sin(i/10000^{2m/d}),    \\
   \textbf{P}(i,2m+1) & = \cos(i/10000^{2m/d}),
  \end{aligned}
\end{equation}
where each dimension of positional encoding corresponds to a sinusoid, and the wavelengths form a geometric progression from $2\pi$ to $10000 \cdot 2 \pi$. Denote $\textbf{P}(i+k)$ and $\textbf{P}(i)$ as encoding vectors for the $(i+k)^{th}$ and $i^{th}$ node, for any fixed offset $k$, $\textbf{P}(i+k)$ can be represented as a linear function of $\textbf{P}(i)$, which means that the positional encoding function can capture the relative temporal relationships of nodes. Thus the initial feature $\mathbf{z}^0_i$ can be obtained as
\begin{equation}
\mathbf{z}^0_i = [\textbf{P}(i,0), \cdots, \textbf{P}(i,d-1)].
\end{equation}

\subsubsection{Feature Pre-Training}

In many static graph clustering methods \cite{xie2016unsupervised, bo2020structural, tu2021deep, liu2022DCRN}, they usually use pre-training technology to generate a priori clustering distribution through node embeddings obtained from pre-trained modules to guide model optimization. The selection of pre-trained modules is often those classic and verified methods, such as AE \cite{DBLP:journals/corr/KingmaW13}, GAE \cite{kipf2016variational}, Deepwalk \cite{perozzi2014deepwalk}, node2vec \cite{grover2016node2vec}, etc.

Since many temporal graph datasets lack original features, using pre-training technology to generate node embeddings as features is also a feasible way, i,e.,
\begin{equation} 
    \mathbf{Z}^0 = \mathcal{F}^{\operatorname{\emph{P}}}(\mathcal{G}).
\end{equation}
$\mathbf{Z}^0$ denotes the initial features (pre-trained embeddings), which can be used in training. $\mathcal{F}^{\operatorname{\emph{P}}}$ denotes the free pre-training module. In the experiment, we select node2vec because it can flexibly adjust the random walk strategy.

\subsection{Training}

As mentioned above, our framework can be deployed on almost any temporal graph methods. For these baselines, we simply use $\mathcal{L}_{\operatorname{\emph{model}}}$ to refer them in training. Here we only discuss the improved classical clustering technologies used on temporal graphs. Note that in temporal graph learning, data will be divided into multiple batches for training. Therefore, the loss functions below need to be improved to adapt this, which is more reflected in code changes.

\subsubsection{Classical Modules}

We first summarize classic loss functions commonly used in graph clustering.

\textbf{Feature Reconstruction.} The most common loss is the reconstruction of initial features $\mathbf{X}$ with node embeddings $\mathbf{Z}$, which encourages $\mathbf{Z}$ to absorb richer information while retaining the original important data, i.e.,
\begin{equation} 
    \mathcal{L}_X = \sum_{i \in \mathcal{B}} \Vert \mathbf{z}_i - \mathbf{x}_i \Vert^2_2.
\end{equation}

\textbf{Distribution Alignment.} Clustering distribution alignment is a common technology in deep clustering methods. Such idea comes from student t-distribution \cite{van2008visualizing}. By constructing a prior distribution, the training distribution is constrained to be aligned with prior, thereby guiding its updating. This relies heavily on the reliability of the prior distribution, which echoes the feature pre-training mentioned above. A reliable prior distribution $\mathbf{P}_{ik}$ between the $i$-th node and the $k$-th clusters can be calclulated as
\begin{equation}
    \mathbf{P}_{ik} = \frac{\mathbf{Q}_{ik}^2/\sum_{i^{\prime}} \mathbf{Q}_{i^{\prime}k}}{\sum_{k^{\prime} = 1}^K [\mathbf{Q}_{ik^{\prime}}^2/\sum_{i'} \mathbf{Q}_{i^{\prime}k^{\prime}}] },
\end{equation}
where
\begin{equation}
    \mathbf{Q}_{ik} = \frac{(1 + \Vert \mathbf{z}_i - \mathbf{c}_k \Vert ^2)^{-\frac{1}{2}}}{\sum_{k^{\prime} = 1}^K (1 + \Vert \mathbf{z}_i - \mathbf{c}_{k^{\prime}} \Vert ^2)^{-\frac{1}{2}}}.
\end{equation}

The initial $\mathbf{Q}$ is calculated using features $\mathbf{X}$ and cluster centers $\mathbf{C}$, and will subsequently change as node embeddings $\mathbf{Z}$ (replace $\mathbf{X}$) are updated.
$\mathbf{P}$ is the sharpened version of $\mathbf{Q}$, resulting in a more distinct node attribution. In subsequent optimization, $\mathbf{Q}$ will continue to change with training, while $\mathbf{P}$ remains unchanged to guide the learning of $\mathbf{Q}$. Then we introduce the Kullback-Leibler (KL) divergence to evaluate these two distributions as
\begin{equation}
    \mathcal{L}_D = \mathbf{KL}(\mathbf{P} \Vert \mathbf{Q}) = \sum_{i \in \mathcal{B}}\sum_{k=1}^K \mathbf{P}_{ik}\log \frac{\mathbf{P}_{ik}}{\mathbf{Q}_{ik}}.
\end{equation}

\textbf{Contrastive Learning.} This is a common technology in many self-supervised fields. The loss function constructed by graph clustering includes two aspects, namely nodes and cluster centers. The purpose of self-supervised calibration is to shorten the distance between the sample and itself and increase the distance with other samples, so as to achieve the uniqueness of each sample, i.e.,
\begin{equation} 
    \mathcal{L}_{C} = \sum_{i \in \mathcal{B}} \mathcal{L}_{\operatorname{\emph{node}}} + \mathcal{L}_{\operatorname{\emph{cluster}}},
\end{equation}
where
\begin{equation} 
    \mathcal{L}_{\operatorname{\emph{node}}} =  \frac{-\log \Vert \mathbf{z}_i - \mathbf{z}_i \Vert^2_2}{-\log \Vert \mathbf{z}_i - \mathbf{z}_i \Vert^2_2 - \log \sum_{n} \Vert \mathbf{z}_i - \mathbf{z}_n \Vert^2_2},
\end{equation}
and
\begin{equation} 
\mathcal{L}_{\operatorname{\emph{cluster}}} = -\log \frac{1}{K} \sum_{k = 1}^K \frac{e^{[\cos(\mathbf{c}_k, \mathbf{c}_k)/\tau]}}{\sum_{k^{\prime} = 1}^K e^{[ \cos(\mathbf{c}_k, \mathbf{c}_{k^{\prime}})/\tau]}}.
\end{equation}

\subsubsection{Advanced Modules}

In addition to the classic modules, there are some novel modules based on the pattern of batch processing.

\textbf{Cross-Batch Calibration.} In temporal graph learning, batch data is the most common form, which means that models need to be optimized and updated in each batch. In order to ensure the consistency of cluster centers in different batches, we propose a cross-batch alignment loss to bring the cluster centers of two batches closer, i.e.,
\begin{equation} 
    \mathcal{L}_B = \Vert \mathbf{C}^b - \mathbf{C}^{b-1} \Vert^2_F.
\end{equation}
Note that different batches have different calculated cluster distributions because they contain different interactions. However, for the whole graph, although the cluster centers will change, it is difficult to change drastically between closely connected batches. Therefore, we set such loss to ensure that the evolution of the cluster centers is gradual.

\textbf{Cluster Scaling.} It is a novel concept from Dink-Net \cite{liu2023dink} that explains the relationship between nodes and cluster centers by the Big Bang. When the universe explodes, different stars (cluster centers) move away from each other, i.e., dilation. At the same time, the planets (nodes) around them move closer to these stars, i.e., shrink. Based on this idea, the cluster scaling loss can be calculated as
\begin{equation} 
    \mathcal{L}_S = \sum_{i \in \mathcal{B}} \mathcal{L}_{\operatorname{\emph{dilation}}} + \mathcal{L}_{\operatorname{\emph{shrink}}} + \Vert \mathbf{C} \Vert^2_F,
\end{equation}
where
\begin{equation} 
    \mathcal{L}_{\operatorname{\emph{dilation}}} = \frac{-1}{(K - 1) K} \sum_{k=1}^{K} \sum_{k^{\prime} \neq k, k^{\prime} = 1}^{K} \Vert \mathbf{c}_k - \mathbf{c}_{k^{\prime}} \Vert^2_2,
\end{equation}
and
\begin{equation} 
    \mathcal{L}_{\operatorname{\emph{shrink}}} = \frac{1}{K}  \sum_{k=1}^{K} \Vert \mathbf{z}_i - \mathbf{c}_k \Vert^2_2.
\end{equation}

\subsubsection{Loss Function}

The total loss function can be formulated as
\begin{equation} 
    \mathcal{L} = \mathcal{L}_{\operatorname{\emph{model}}} + \mathcal{L}_{\operatorname{\emph{module}}},
\end{equation}
where $\mathcal{L}_{\operatorname{\emph{model}}}$ denotes the baseline method loss, and $\mathcal{L}_{\operatorname{\emph{module}}}$ denotes the improved clustering loss functions mentioned above. Such functions need not to be deployed all, which can be flexibly selected and matched according to different cases.

\subsection{Clustering}

Since K-means is so successful, almost all existing clustering methods use K-means to complete the final clustering evaluation. Graph clustering methods can be divided into two categories according to whether use K-means.

\subsubsection{One-Step Graph Clustering}

One-step graph clustering is an ideal way that has not yet been studied \cite{liu2021one, lwx2024}. It means that during training, the method learns a potential distribution \cite{van2020scan, tsai2020mice, huang2020deep}. When training is complete, the clustering result is directly output from such potential distribution without K-means. That is, from data to clustering results. Although one-step clustering has been explored in some methods \cite{li2021contrastive, shen2021you, li2022twin}, it has hardly been explored in graph clustering. This means designing new model structures to replace the use of K-means, which is beyond our scope.

\subsubsection{Two-Step Graph Clustering}

Existing methods are almost all two-step graph clustering. The sample representations obtained during training will be sent to K-means for clustering. That is, from data to representations, then to K-means. \emph{Our framework and experiments are all based on the two-step clustering construction, which is consistent with existing methods.}

\subsection{Complexity Discussion}

Temporal graph methods are trained in batches, and the division of batches is independent of the number of nodes ($\mathcal{N}$), but rather based on the length of the interaction sequence ($\mathcal{E}$), i.e., the total number of interactions. This implies that the main complexity, in terms of both time and space, in temporal graph clustering is $O(\mathcal{E})$, rather than $O(\mathcal{N}^2)$, as observed in static graph clustering. The advantage of temporal graph clustering lies in the fact that it does not require accessing the entire adjacency matrix. Specifically:

(1) In most cases, $\mathcal{E}$ is much smaller than $\mathcal{N}^2$ because the upper bound of $\mathcal{E}$ is $\mathcal{N}^2$ (as a fully connected graph). Thus $O(\mathcal{N}^2) \gg O(\mathcal{E})$ holds true in the majority of scenarios.

(2) In certain cases, $O(\mathcal{N}^2) < O(\mathcal{E})$, indicating multiple interactions between a substantial number of node pairs. This highlights the superiority of dynamic interaction sequences over the adjacency matrix, as the latter compresses multiple interactions into a single edge, resulting in significant loss of information (This will be verified in the experiment).

(3) The aforementioned discussion applies not only to time complexity but also to space complexity. Since the interaction sequence is chronologically arranged, it can be partitioned into multiple batches of varying sizes. The maximum batch size is primarily determined by the available memory of the deployed platform and can be dynamically adjusted to accommodate memory constraints.

In our experiments, we consider various types of datasets, encompassing dissimilar node degrees and scales, as elaborated in the next section.

\section{BenchTGC Datasets}

In this section, we discuss another challenge (\textbf{Inapplicable datasets}), and solve it with \textbf{BenchTGC Datasets}.

\subsection{Public Datasets}

As previously indicated, the advancement of deep temporal graph clustering is currently hindered by a notable challenge: the scarcity of appropriate and dependable public datasets for evaluating clustering performance \cite{pareja2020evolvegcn}. While temporal graph clustering can be trained in an unsupervised setting, node labels are indispensable for evaluating experimental performance. Unfortunately, a majority of temporal graph datasets encounter one or more of the following issues:


(1) Due to the prevalence of public temporal graph datasets that lack labels or have limited label diversity (consisting mainly of 0 and a few 1), many temporal graph methods have been developed specifically for the link prediction task. These methods do not rely on node labels, thus do not consider node clustering. Instead, they focus on predicting the activity of nodes at specific moments or future. Given the binary nature of these datasets with only two label types, they are better suited for binary classification tasks rather than multi-class classification, rendering them less suitable for node clustering purposes.

(2) In certain datasets, node labels may not adequately align with the inherent characteristics of the nodes. For instance, when considering product ratings provided by users (typically ranging from 1 to 5) as labels, it can hardly claim that these label of ratings are more relevant to the products than the product category labels. Consequently, this misalignment can lead to subpar clustering performance on such datasets. In unsupervised scenarios, node clustering tends to group products into different categories (e.g., toys or foods) rather than based on their ratings.

(3) Only a limited number of datasets are well-suited for temporal graph clustering, and even these datasets tend to be relatively small in scale.

\begin{figure}[t]
    \centering
    \includegraphics[width=0.48\textwidth]{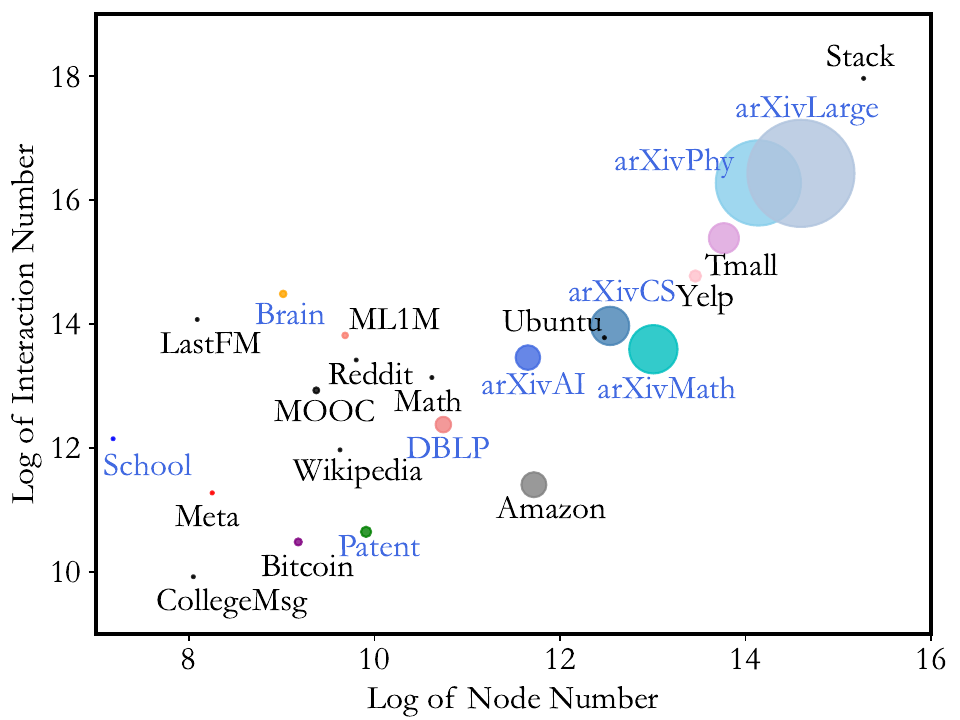}
    \caption{Scatter plots showcasing various dataset scales. The X-axis corresponds to the number of nodes, while the Y-axis represents the interaction count. To account for the substantial variations across datasets, we present the data size using a logarithmic scale. The scatter size corresponds to the quantity of node labels accessible for each dataset.}
    \label{scatter}
\end{figure}

\begin{table}[t]
    \caption{Comparison of public temporal graph datasets. The first three parts of the dataset are unlabeled, clustering unavailable, and clustering available, respectively. The last part of the datasets are the datasets we developed for TGC task, i.e., BenchTGC Datasets.}
    \centering
    \resizebox{0.98\linewidth}{!}{
        \begin{tabular}{c|rrrrr}
            \toprule[1.5pt]
            Dataset    & Nodes     & Interactions & K & Labels & Timestamps   \\
            \midrule[0.5pt]
            CollegeMsg   & 1,899     & 20,296       & N/A     & N/A & 193    \\
            LastFM   & 1,980     & 1,293,103    & N/A     & N/A &  Seconds      \\
            MOOC       & 7,144     & 411,749      & N/A     & N/A &  Seconds      \\
            Wikipedia      & 9,228     & 157,474      & N/A     & N/A & Seconds  \\
            Reddit     & 10,985    & 672,447      & N/A     & N/A  &  Seconds    \\
            Math   & 24,818   & 506,550      & N/A     & N/A &  2,350      \\
            Ubuntu     & 159,316   & 964,437      & N/A     & N/A &  2,613      \\
            Stack   & 2,601,977   & 63,497,050      & N/A     & N/A &  2,774      \\
            \midrule[0.5pt]
            Meta    & 2,327     & 78,600       & 4/5/6    & 1,166 &  366  \\
            Bitcoin    & 5,881     & 35,592       & 21    & 5,858 &  27,487  \\
            ML1M       & 9,746     & 1,000,209    & 5     & 3,706  &  458,455 \\
            Amazon     & 74,526    & 89,689       & 5     & 72,098 & 5,804  \\
            Yelp     & 424,450   & 2,610,143    & 5     & 15,154 & 153  \\
            Tmall     & 577,314   & 4,807,545    & 10    & 104,410 & 186 \\
            \midrule[0.5pt]
            Brain      & 5,000     & 1,955,488    & 10    & 5,000  &  12 \\
            Patent    & 12,214    & 41,916       & 6     & 12,214 &  891 \\
            DBLP      & 28,085    & 236,894      & 10    & 28,085 & 27  \\
            \midrule[0.5pt]
            School      & 327    & 188,508      & 9    & 327 & 7,375  \\
            arXivAI    & 69,854    & 699,206      & 5     & 69,854  & 27 \\
            arXivCS    & 169,343   & 1,166,243    & 40    & 169,343 & 29 \\
            arXivMath  & 270,013   & 799,745      & 31    & 270,013 & 31 \\
            arXivPhy   & 837,212   & 11,733,619   & 53    & 837,212 & 41 \\
            arXivLarge & 1,324,064 & 13,701,428   & 172   & 1,324,064 & 41 \\
            \bottomrule[1.5pt]
        \end{tabular}}
    \label{datasets}
\end{table}

To investigate these issues in more depth, we conduct a comparative analysis of multiple public temporal graph datasets. Figure \ref{scatter} provides a visual representation of the different dataset scales, which have been logarithmized for enhanced distinguishability. This logarithmic transformation ensures that the datasets are clearly discernible. Had we used the original data magnitudes, the majority of datasets would have been concentrated in the lower left quadrant, making it challenging to differentiate between them. Table \ref{datasets} gives the details of these datasets.

In light of the aforementioned issues, the public available datasets can be categorized into three groups:

(1) \textbf{Unlabeled datasets:} CollegeMsg \cite{panzarasa2009patterns}, LastFM \cite{kumar2019predicting}, MOOC \cite{kumar2019predicting}, Wikipedia \cite{kumar2019predicting}, Reddit \cite{kumar2019predicting}, Math \cite{paranjape2017motifs}, Ubuntu, and Stack. Note that \emph{Math}, \emph{Ubuntu}, and \emph{Stack} are three temporal interaction graphs derived from the Stack Exchange websites ``Math Overflow'', ``Ask Ubuntu'', and ``Stack Overflow'', respectively.

(2) \textbf{Clustering unavailable datasets:} Meta, Bitcoin \cite{kumar2016edge}, ML1M \cite{li2020time}, Amazon \cite{ni2019justifying}, Yelp \cite{zuo2018embedding}, and Tmall \cite{zuo2018embedding}. \emph{Meta} provides blockchain financial transactions within the Open Metaverse\footnote{\url{https://www.openmv.org/}} website, and it was obtained from the Kaggle website. \emph{Amazon} \cite{ni2019justifying} is a dataset representing magazine subscription graphs from the Amazon platform.


(3) \textbf{Clustering available datasets:} Brain \cite{preti2017dynamic}, Patent \cite{hall2001nber}, and DBLP \cite{zuo2018embedding}. \emph{Brain} represents the connectivity graph of brain tissue in humans.

\subsection{Discussion}
\label{case study section}

As above, we divide the public datasets into three types: unlabeled, clustering unavailable, and clustering available. 

For unlabeled datasets, although there are internal indicators (DBI / CHI) can be used to measure their clustering performance, these indicators often focus on the clarity of label assignment rather than accuracy. In other words, internal indicators are actually unable to measure clustering performance in real scenarios. This has also been discussed and confirmed in many works and will not be repeated here.

What we really want to highlight is the second type of datasets, where clustering unavailable datasets. Although these datasets are labeled, they are not suitable for clustering. There are two main metrics for clustering, NMI and ARI, with very low results on these datasets. Furthermore, different label selections will significantly affect the clustering performance. Based on the above judgment, we conduct a case study to discuss them.

\begin{table}
\caption{Case study on Meta dataset. We use different taxonomies to construct labels. Regardless of the labels, the NMI and ARI metrics of these methods are very poor.}
    \centering
    \begin{tabular}{cc|ccc}
    \toprule[1.5pt]
         Taxonomy& Metric & node2vec & SDCN & HTNE \\
         \midrule[0.5pt]
         ~& ACC & 29.37 & 29.45 & 26.87\\
         PUR& NMI & 0.19 & 0.31 & 0.16\\
         ($K$=4)& ARI & -0.01 & 0.18 & -0.15\\
         ~& F1 & 26.67 & 26.44 & 25.15\\
         \midrule[0.5pt]
         ~& ACC & 23.42 & 28.07 & 24.37\\
         TRA& NMI & 0.55 & 0.47 & 0.42\\
         ($K$=5)& ARI & -0.02 & -0.15 & -0.25\\
         ~& F1 & 23.29 & 21.21 & 19.60\\
         \midrule[0.5pt]
         ~& ACC & 22.65 & 24.63 & 24.72\\
         ANO& NMI & 0.65 & 0.65 & 0.62\\
         ($K$=6)& ARI & 0.09 & 0.91 & 0.13\\
         ~& F1 & 17.87 & 18.66 & 18.23\\
         \bottomrule[1.5pt]
    \end{tabular}
    \label{meta}
\end{table}

In Table \ref{meta}, we report the performance of multiple taxonomies on Meta dataset. Meta is a financial transaction graph, which we marked as clustering unavailable. In its origin data, each transaction has multiple levels of categories, such as IP, location, risk score, age, etc. Due to the different category levels, the label types are also different. We select three representative levels as node labels, i.e., purchase pattern (PUR, $K$=4), transaction type (TRA, $K$=5), anomaly (ANO, $K$=6). We select three baseline methods, where node2vec is a classical method, SDCN is a static graph clustering method, and HTNE is a temporal graph learning method, which will be introduced in the experiment section.

We put this experiment in the dataset section early because it is closely related to this discussion. As shown in the table, we can find that: (1) when label changes, the clustering performance will change accordingly, which shows the importance of label selection to the experimental results, and (2) regardless of the selection of labels, the NMI and ARI indicators of all methods are very poor. 

In particular, compared to ACC and F1-score, which may be calculated by merging or relabeling clusters, NMI and ARI are more rigorous measures of the match between the clustering results and the true labels. In this case, if the correspondence between the clustered labels and the true labels is not clear, NMI and ARI may show low values. This means that if the data points within a cluster center are very close together, while the data points between cluster centers are widely distributed, it may cause the ACC and F1-score to look high, while NMI and ARI are low because they take into account the separation between clusters. In this way, it is difficult to measure the quality of clustering using such labels, and it is impossible to observe the improvements.

In this case, the first and second types of datasets are actually unusable for TGC, and the third type of datasets are relatively small, so we further developed more datasets.

\subsection{Developed Datasets}

\begin{table}[t]
    \caption{Label meanings in the developed datasets. Labels in School dataset are composed of different classes, and labels in the set of arXiv datasets come from different research fields defined by the arXiv website.}
    \centering
        \resizebox{\linewidth}{!}{
        \begin{tabular}{c c}
            \toprule[1.5pt]
            Dataset      & Label Meaning    \\
            \midrule[0.5pt]
            School         & \makecell[c]{Mathematics and physics (\textbf{MP, MP*1, MP*2}), \\ Physics and chemistry (\textbf{PC, PC*}), \\ Engineering (\textbf{PSI}), Biology (\textbf{2BIO1, 2BIO2, 2BIO3})}\\ 
            \midrule[0.5pt]
            arXivAI         & \makecell[c]{Artificial Intelligence (\textbf{arxiv cs ai}), Machine Learning (\textbf{arxiv cs lg}), \\ Computation and Language (\textbf{arxiv cs cl}), \\ Computer Vision and Pattern Recognition (\textbf{arxiv cs cv}), \\ Neural and Evolutionary Computing (\textbf{arxiv cs ne})} \\  
            \midrule[0.5pt]
            arXivCS & \makecell[c]{\textbf{arxiv cs:} na, mm, lo, cy, cr, dc, hc, ce, ni, cc, ai, \\ ma, gl, ne, sc, ar, cv, gr,et, sy, cg, oh, pl, se, lg, sd, \\ si, ro, it, pf, cl, ir, ms, fl, ds, os, gt, db, dl, dm} \\
            \midrule[0.5pt]
            arXivMath & \makecell[c]{\textbf{arxiv math:} ac, ag, ap, at, ca, co, ct, cv, dg, ds, fa, gm, gn, gr,\\ gt, ho, kt, lo, mg, na, nt, oa, oc, ph, pr, qa, ra, rt, sg, sp, st}\\
            \midrule[0.5pt]
            arXivPhy & \makecell[c]{\textbf{arxiv:} gr qc, hep ex, hep lat, hep ph, hep th, math ph, nucl ex, \\nucl th, quant ph, \textbf{arxiv astro ph:} \emph{self}, co, ep, ga, he, im, sr,\\ \textbf{arxiv nlin}: ao, cd, cg, ps, si, \textbf{arxiv cond mat:} \emph{self}, dis nn, mes hall, \\mtrl sci, other, quant gas, soft, stat mech, str el, supr con, \\ \textbf{arxiv physics}: acc ph, ao ph, app ph, atm clus, atom ph, \\bio ph, chem ph, class ph, comp ph, data an, ed ph, \\flu dyn, gen ph, geo ph, hist ph, ins det, med ph, \\optics, plasm ph, pop ph, soc ph, space ph}\\
            \midrule[0.5pt]
            arXivLarge & \makecell[c]{All \textbf{172} research areas in the fields of Computer Science, Physics, \\ Economics Science, Mathematics, Electrical Engineering and Systems, \\ Quantitative Biology, Quantitative Finance, and Statistics.}\\
            \bottomrule[1.5pt]     
    \end{tabular}}
    \label{fields}
\end{table}

The existence of the above problems makes us urgently need to develop a set of datasets suitable for TGC. In this case, BenchTGC Datasets are born. 

According to Table \ref{fields}, we report the developed datasets. School is a social network dataset, which records high school student interactions in 5 days \cite{mastrandrea2015contact}. In this dataset, there exist students in different classes, and we label them according to the class they belong to, and there are 9 categories. Although this dataset is already has been proposed by statistical researchers, it is rare to see its use in temporal graph learning. Therefore, we normalize and clean it, then construct it as a temporal graph dataset.



The other 5 datasets are a set of arXiv datasets that is extracted from the publicly available ogbn-papers100M dataset, which was curated by OGB \cite{wang2020microsoft}. The largest dataset is called arXivLarge, which contains 1.3 million labeled nodes representing papers and 13 million temporal edges representing citations. It encompasses all 172 subfields listed by arXiv platform\footnote{\url{https://arxiv.org/category_taxonomy}}. The other 4 datasets are subsets of arXivLarge and provide researchers with more flexible options for selecting data scales. Information in these datasets is anonymized. The datasets are easily accessible to researchers, allowing them to analyze the relationships and interactions between papers in various fields of study.

Although the sources of these datasets are public, since they contain a large amount of unlabeled data, we need to do a lot of additional processing:

(1) We extract the node interactions from the original data, identify their respective node labels, and eliminate the unlabeled nodes along with their corresponding interactions (referred to as arXivLarge).

(2) We extracted a subset of edges into a new dataset based on the domain associated with the node labels. Throughout this process, we explore various domain combinations and ultimately select 4 combinations to retain: arXivAI, arXivCS, arXivMath, and arXivPhy.

(3) As the node numbers in the extracted datasets were scattered, we renumber them and update their node labels.

(4) To ensure a fair comparison, we enhance the node features by incorporating position encoding technology in addition to the original features.

\begin{figure*}[t]
    \centering
    \begin{minipage}[t]{0.16\textwidth}
        \includegraphics[width=1\textwidth]{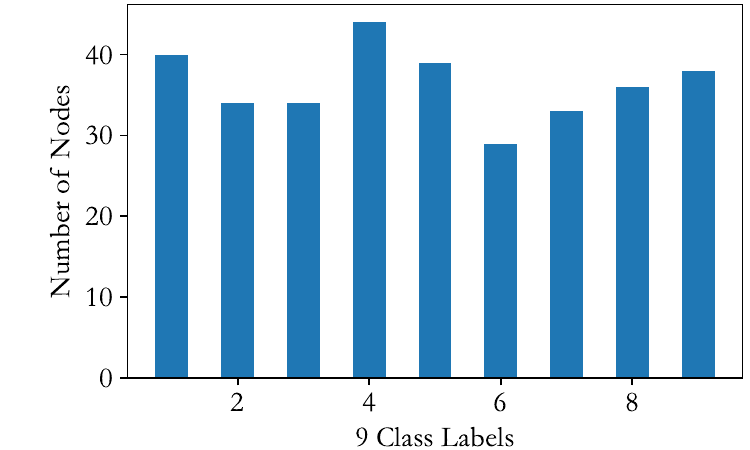}
        \centerline{\quad(a) School}
    \end{minipage}%
    \hspace{0.5mm}
    \begin{minipage}[t]{0.16\textwidth}
        \includegraphics[width=1\textwidth]{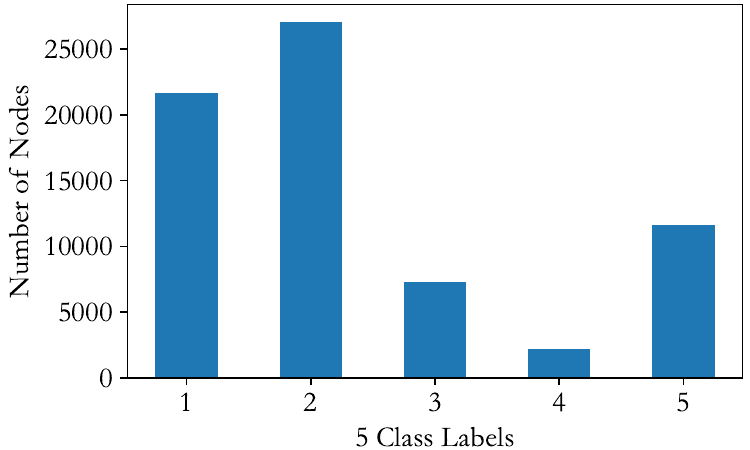}
        \centerline{\quad(b) arXivAI}
    \end{minipage}%
    \hspace{0.5mm}
    \begin{minipage}[t]{0.16\textwidth}
        \includegraphics[width=1\textwidth]{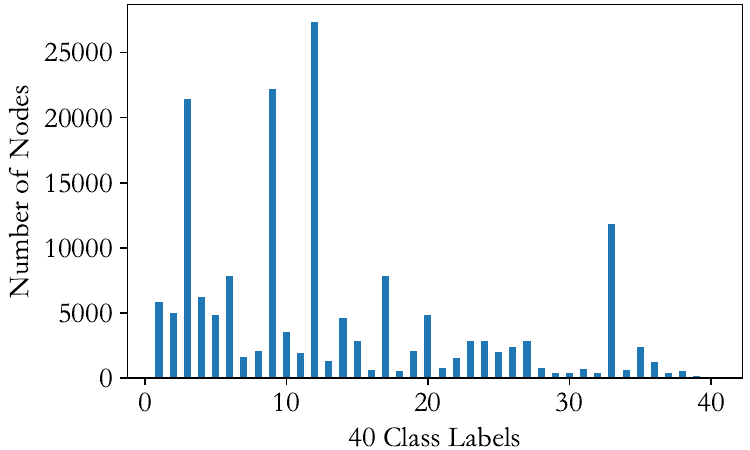}
        \centerline{\quad(c) arXivCS}
    \end{minipage}%
    \hspace{0.5mm}
    \begin{minipage}[t]{0.16\textwidth}
        \includegraphics[width=1\textwidth]{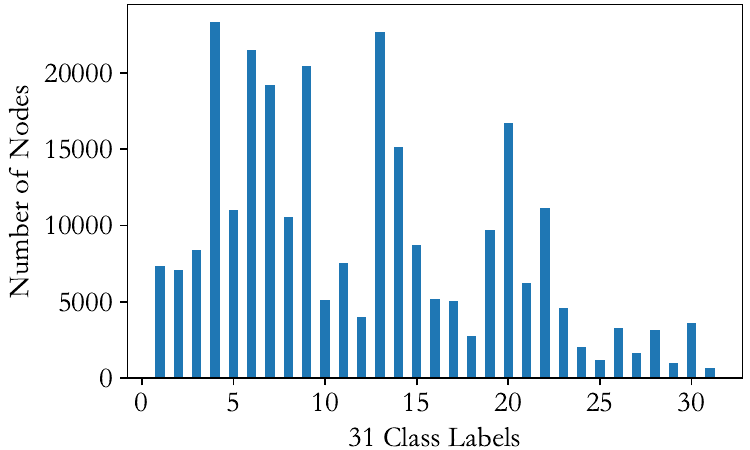}
        \centerline{\quad(d) arXivMath}
    \end{minipage}%
    \hspace{0.5mm}
        \begin{minipage}[t]{0.16\textwidth}
        \includegraphics[width=1\textwidth]{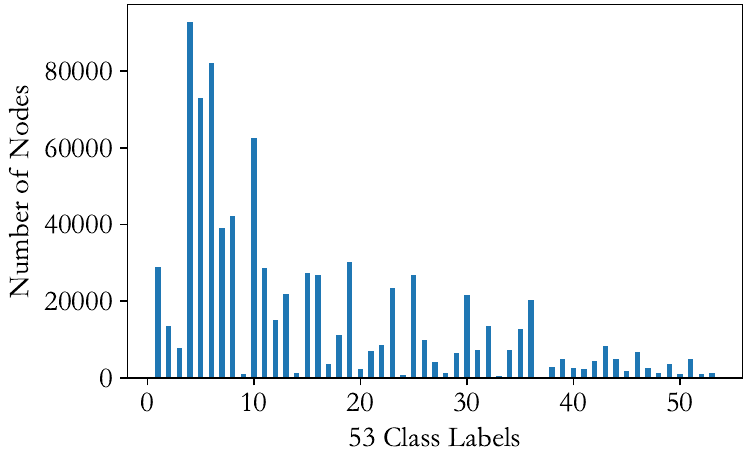}
        \centerline{\quad(e) arXivPhy}
    \end{minipage}%
    \hspace{0.5mm}
        \begin{minipage}[t]{0.16\textwidth}
        \includegraphics[width=1\textwidth]{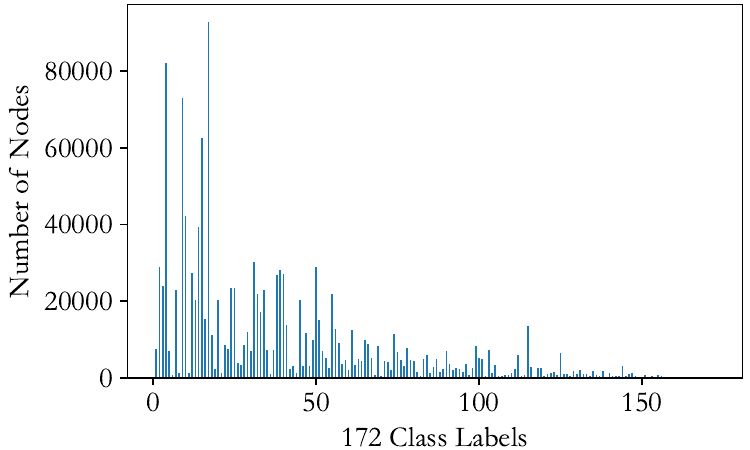}
        \centerline{\quad(f) arXivLarge}
    \end{minipage}%
    \caption{Class Distributions of our 6 developed datasets. Each bar represents a cluster, and the bar height denotes the node number it contains.}
    \label{dis}
\end{figure*}

(5) We perform a comprehensive analysis of the class distribution for these datasets, as illustrated in Figure \ref{dis}. This distribution aligns with the inherent diversity of researchers in various fields within real-world academic research. Furthermore, it facilitates the investigation of category imbalance problems by researchers.

(6) We consolidate and compress these datasets to enable public access and download. Additionally, we provide a baseline model specifically designed for the Temporal Graph Clustering (TGC) task. This facilitates the utilization of these datasets to obtain benchmark results with ease.

(7) It is worth mentioning that in addition to developing 6 new datasets, we also screened out a small number of 3 available datasets from 100+ public datasets, namely DBLP, Brain, and Patent. This work of screening datasets means a greater workload than developing datasets.

\begin{table*}[t]
    \centering
    \caption{Dataset statistics of Data4TGC.}
    \label{Dataset statistics}
    \begin{threeparttable}
            \begin{tabular}{c|c c c c c c c c c}
                \toprule[1.5pt]
                Datasets& Nodes& Interactions& Edges& Complexity& Timestamps& $K$& Degree& MinI& MaxI\\
                \midrule[0.5pt]
                DBLP& 28,085& 236,894& 162,441& $\mathcal{N}^2 \gg \mathcal{E}$& 27& 10& 16.87& 1& 955\\
                Brain& 5,000& 1,955,488& 1,751,910& $\mathcal{N}^2 > \mathcal{E}$& 12& 10& 782& 484& 1,456\\
                Patent& 12,214& 41,916& 41,915& $\mathcal{N}^2 \gg \mathcal{E}$& 891& 6& 6.86& 1& 789\\
                School& 327& 188,508& 5,802& $\mathcal{N}^2 < \mathcal{E}$& 7,375& 9& 1153& 7& 4,647\\
                arXivAI& 69,854& 699,206& 699,198& $\mathcal{N}^2 \gg \mathcal{E}$& 27& 5& 20.02& 1& 11,594\\
                arXivCS& 169,343& 1,166,243& 1,166,237& $\mathcal{N}^2 \gg \mathcal{E}$& 29& 40& 13.77& 1& 13,161\\
                arXivMath& 270,013& 799,745& 799,745& $\mathcal{N}^2 \gg \mathcal{E}$& 31& 31& 5.92& 1& 999\\
                arXivPhy& 837,212& 11,733,619& 11,733,614& $\mathcal{N}^2 \gg \mathcal{E}$& 41& 53& 28.03& 1& 7,594\\
                arXivLarge& 1,324,064& 13,701,428& 13,701,425& $\mathcal{N}^2 \gg \mathcal{E}$& 41& 172& 20.70& 1& 12,586\\
                \bottomrule[1.5pt]
        \end{tabular}
    \end{threeparttable}
\end{table*}

\subsection{Data4TGC}

Ultimately, merging the existing publicly available datasets with BenchTGC Datasets yields the 9 datasets currently applicable to TGC task, which we call Data4TGC and open-source at \url{https://github.com/MGitHubL/Data4TGC}.

In the experiment, we utilize these datasets for TGC evaluation. We report their statistics in Table \ref{Dataset statistics}. \emph{Nodes} and \emph{Interactions} represent node number and interaction number, respectively. There is an additional \emph{Edges} used to refer to the edge number in the adjacency matrix when we transfer temporal graphs into static graphs for static graph clustering methods. In this case, no matter how many times two nodes interact, they are only considered one edge. \emph{Complexity} represents the main complexity comparison between static and temporal graph clustering, \emph{Timestamps} denotes interaction time, $K$ means the cluster number, and \emph{Degree} represents node average degree. \emph{MinI} and \emph{MaxI} represent the minimum and maximum interaction times of nodes, respectively.

\section{Experiments}

In conjunction with the experimental results, we will validate and discuss the proposed BenchTGC. Here we ask several important questions and will answer these questions in the experiment:

\textbf{Q1:} \emph{Can BenchTGC Framework bring improvements?}

\textbf{Q2:} \emph{Does BenchTGC Datasets make sense?}

\textbf{Q3:} \emph{How does TGC reduce memory pressure?}

\textbf{Q4:} \emph{How is the flexibility of TGC reflected?}

\textbf{Q5:} \emph{How to use the BenchTGC Framework?}


\subsection{Baselines}
In the experiment, we select 16 classical and SOTA methods for comparison. Such methods contain three types: classical methods, static graph clustering methods, and temporal graph learning methods\footnote{Since TGC task is a new definition, there is currently no temporal graph clustering method, so we compare both static graph clustering methods and temporal graph learning methods.}.

(1) Classical methods: \textbf{K-means} \cite{macqueen1967some} is the classical clustering algorithm which have been used by many methods for performance evaluation. In addition, K-means can also be seen as a method to cluster the origin features directly. AutoEncoder \cite{DBLP:journals/corr/KingmaW13} (known as \textbf{AE}) is a classical method that proposes the encoder-decoder architecture. \textbf{DeepWalk} \cite{perozzi2014deepwalk} is the pioneering work that places emphasis on utilizing random walks on graphs. \textbf{node2vec} \cite{grover2016node2vec} can flexibly adjust the balance between breadth-first and depth-first strategies in random walk. \textbf{GAE} \cite{kipf2016variational} improves variational autoencoders to apply on graph data.

(2) Static graph clustering methods: \textbf{DAEGC} \cite{wang2019attributed} encodes the structure and features into a compact representation by utilizing an attention network. \textbf{MVGRL} \cite{hassani2020contrastive} generates node representations by contrasting structural views of graphs. \textbf{SDCN} \cite{bo2020structural} integrates structural information into sample clustering, and \textbf{SDCNQ} is it variant. \textbf{DFCN} \cite{tu2021deep} utilizes two sub-networks to independently process augmented graphs. \textbf{SCGC} \cite{liu2023simple} designs a straightforward data argumentation network for rapid node clustering.

(3) Temporal graph learning methods: \textbf{HTNE} \cite{zuo2018embedding} utilizes the Hawkes process to model the neighborhood influence for the future event prediction. \textbf{JODIE} \cite{kumar2019predicting} can predict the future node embeddings with the user-item RNN design. \textbf{MNCI} \cite{MNCI_ML_SIGIR} considers the community influence and neighborhood influence in temporal graph learning. \textbf{TREND} \cite{wen2022trend} designs a dynamic graph neural network to combine multi-hop node embeddings. \textbf{S2T} \cite{S2T_ML_TNNLS} aligns the temporal and structural conditional intensities in node interactions.

We deploy the BenchTGC framework on such temporal graph learning methods\footnote{Without affecting the performance of these baseline methods, we adjust the codes of some methods to improve their efficiency.}, and these improved methods can be denote as ``IHTNE'' (i.e., Improved HTNE), ``IJODIE'', ``IMNCI'', ``ITREND'', and ``IS2T'', respectively. 

\subsection{Experimental Settings}

For clustering performance evaluation, we utilize 4 common metrics: ACC (Accuracy), NMI (Normalized Mutual Information), ARI (Adjust Rand Index), and F1 score. Such metrics are external metrics, that is, we need ground-truth labels to calculate these metrics for evaluation.


For all comparison methods, we use their default parameters. Our BenchTGC Framework improves on 5 temporal graph methods, and some of the same hyper-parameters values in these methods are still different. We basically keep all the default values of these methods, although they may be at different learning rates or epochs. The only hyper-parameter value we unified is the embedding size, which we defaulted to 128 (a commonly used value) to achieve alignment between different stages and different methods.

In the experiment, we set node clustering as the main experiment to compare performance. In addition, we also explore other characteristics of TGC, including case study (in Section \ref{case study section}), GPU memory analysis, time and space analysis, ablation study, and visualizations, etc.

Our proposed BenchTGC framework is implemented with PyTorch, and all experiments are running on one NVIDIA RTX 4090 (24 GB) GPU, 90 GB RAM, and 2.10GHz Intel Xeon Platinum 8352V CPU.

\subsection{Node Clustering Performance}

\begin{table*}[t]
    \centering
        \caption{Clustering performance in all datasets. The best results are highlighted in \textbf{bold}, and the second best results are highlighted in \underline{underline}. When utilizing the BenchTGC Framework on HTNE, we refer to it as ``IH'', which stands for ``Improved HTNE''. If BenchTGC Framework yields improvements over the baseline method, we denote that result as grayed out. In cases where a method encounters the out-of-memory problem, it is recorded as OOM (Out-Of-Memory). For brevity, we abbreviate ``Deepwalk'' and ``node2vec'' as ``DW'' and ``N2V'', respectively. Each result was executed 5 times, and the average value is reported. To save space, the std results is not given.}
        \renewcommand\arraystretch{1.1}
    \renewcommand\tabcolsep{0.1pt}
    \resizebox{1\textwidth}{!}{
    \begin{tabular}{C{1.4cm} C{1.0cm} | C{1.1cm} C{0.7cm} C{0.8cm} C{0.8cm} C{0.8cm} C{1.1cm} C{1.1cm} C{0.9cm} C{1.1cm} C{0.9cm} C{0.9cm} | C{1.0cm} C{0.8cm} C{0.9cm} C{0.8cm} C{0.9cm} C{0.8cm} C{1.1cm} C{0.8cm} C{0.8cm} C{0.8cm}}
    \toprule[1.5pt]
        Dataset & Metric & Kmeans & AE & DW & N2V & GAE & DAEGC & MVGRL & SDCN & SDCNQ & DFCN & SCGC & HTNE & \cellcolor{gray!10}IH & JODIE & \cellcolor{gray!10}IJ & MNCI & \cellcolor{gray!10}IM & TREND & \cellcolor{gray!10}IT & S2T & \cellcolor{gray!10}IS \\
        \midrule[0.5pt] 
        \multirow{4}{*}{DBLP}& ACC & 39.17  & 42.16  & 45.07  & 46.31  & 39.31  & \multirow{4}{*}{OOM} & 28.95  & 46.69  & 40.47  & 41.97  & 43.20  & 45.74  & \cellcolor{gray!10}48.71  & 20.79  & \cellcolor{gray!10}46.35  & 45.83  & \cellcolor{gray!10}48.36  & 46.82  & \cellcolor{gray!10}\textbf{55.10}  & 46.21  & \cellcolor{gray!10}\underline{50.88}   \\
        ~ & NMI & 34.50  & 36.71  & 31.46  & 34.87  & 29.75  & ~ & 22.03  & 35.07  & 31.86  & 36.94  & 29.38  & 35.95  & \cellcolor{gray!10}37.26  & 1.70  & \cellcolor{gray!10}34.90  & 36.09  & \cellcolor{gray!10}36.93  & 36.56  & \cellcolor{gray!10}\textbf{43.33}  & 32.47  & \cellcolor{gray!10}\underline{40.09}   \\
        ~ & ARI & 21.91  & 22.54  & 17.89  & 20.40  & 17.17  & ~ & 13.73  & 23.74  & 19.80  & 21.46  & 16.72  & 22.13  & \cellcolor{gray!10}22.93  & 1.64  & \cellcolor{gray!10}20.49  & 22.00  & \cellcolor{gray!10}22.46  & 22.83  & \cellcolor{gray!10}\textbf{31.09}  & 22.32  & \cellcolor{gray!10}\underline{27.51}   \\
        ~ & F1 & 34.79  & 37.84  & 38.56  & 43.35  & 35.04  & ~ & 24.79  & 40.31  & 35.18  & 35.97  & 40.92  & 43.98  & \cellcolor{gray!10}45.03  & 13.23  & \cellcolor{gray!10}43.39  & 43.56  & \cellcolor{gray!10}44.84  & 44.54  & \cellcolor{gray!10}\textbf{49.94}  & 42.51  & \cellcolor{gray!10}\underline{48.05}   \\
        \midrule[0.5pt] 
        \multirow{4}{*}{Brain}& ACC & 43.84  & 43.48  & 41.28  & 43.92  & 31.22  & 42.52  & 15.76  & 42.62  & 43.42  & 42.08  & 42.98  & 43.20  & \cellcolor{gray!10}43.86  & 19.14  & \cellcolor{gray!10}43.52  & 29.60  & \cellcolor{gray!10}\underline{43.96}  & 39.83  & \cellcolor{gray!10}45.90  & 34.32  & \cellcolor{gray!10}\textbf{48.86}   \\
        ~ & NMI & 49.90  & 50.49  & 49.09  & 45.96  & 32.23  & 49.86  & 21.15  & 46.61  & 47.40  & 43.24  & 45.87  & 50.33  & \cellcolor{gray!10}\textbf{51.21}  & 10.50  & \cellcolor{gray!10}50.63  & 32.17  & \cellcolor{gray!10}\underline{50.83}  & 45.64  & \cellcolor{gray!10}48.34  & 39.81  & \cellcolor{gray!10}50.49   \\
        ~ & ARI & 28.81  & 29.78  & 27.89  & 26.08  & 14.97  & 27.47  & 9.77  & 27.93  & 27.69  & 24.62  & 26.11  & 29.26  & \cellcolor{gray!10}\textbf{30.25}  & 5.00  & \cellcolor{gray!10}29.79  & 15.83  & \cellcolor{gray!10}28.86  & 22.82  & \cellcolor{gray!10}28.40  & 21.38  & \cellcolor{gray!10}\underline{29.94}   \\
        ~ & F1 & 43.05  & 43.26  & 42.54  & 46.61  & 34.11  & 43.24  & 13.56  & 41.42  & 37.27  & 42.73  & 45.14  & 43.85  & \cellcolor{gray!10}44.23  & 11.12  & \cellcolor{gray!10}42.64  & 27.41  & \cellcolor{gray!10}43.83  & 33.67  & \cellcolor{gray!10}\underline{48.76} & 32.85  & \cellcolor{gray!10}\textbf{51.42}   \\
        \midrule[0.5pt] 
        \multirow{4}{*}{Patent} & ACC & 28.98  & 30.81  & 38.69  & 40.36  & 39.65  & \underline{46.64}  & 31.13  & 37.28  & 32.76  & 39.23  & 34.40  & 45.07  & \cellcolor{gray!10}\textbf{50.51}  & 30.82  & \cellcolor{gray!10}43.56  & 40.80  & 38.48  & 38.72  & \cellcolor{gray!10}43.02  & 31.68  & \cellcolor{gray!10}42.81   \\
        ~ & NMI & 7.27  & 8.76  & 22.71  & 24.84  & 17.73  & 21.28  & 10.19  & 13.17  & 9.11  & 15.42  & 15.08  & 20.77  & \cellcolor{gray!10}\textbf{34.65}  & 9.55  & \cellcolor{gray!10}25.15  & 17.21  & \cellcolor{gray!10}25.05  & 14.44  & \cellcolor{gray!10}21.89  & 11.42  & \cellcolor{gray!10}\underline{30.28}   \\
        ~ & ARI & 5.73  & 7.43  & 10.32  & 18.95  & 13.61  & 16.74  & 10.26  & 10.12  & 7.84  & 12.24  & 11.75  & 10.69  & \cellcolor{gray!10}\textbf{29.12}  & 7.46  & \cellcolor{gray!10}19.42  & 11.25  & \cellcolor{gray!10}16.50  & 13.45  & \cellcolor{gray!10}15.28  & 8.23  & \cellcolor{gray!10}\underline{21.23}   \\
        ~ & F1 & 24.85  & 26.65  & 31.48  & 34.97  & 30.95  & 32.83  & 18.06  & 31.38  & 28.27  & 30.32  & 29.32  & 28.85  & \cellcolor{gray!10}\textbf{39.96}  & 20.83  & \cellcolor{gray!10}34.45  & 32.96  & \cellcolor{gray!10}33.48  & 28.41  & \cellcolor{gray!10}\underline{39.06}  & 27.34  & \cellcolor{gray!10}33.74   \\
        \midrule[0.5pt] 
        \multirow{4}{*}{School} & ACC & 30.88  & 30.88  & 90.60  & 100.0  & 85.62  & 34.25  & 32.37  & 48.32  & 33.94  & 49.85  & 22.69  & 99.38  & \cellcolor{gray!10}99.69  & 19.88  & \cellcolor{gray!10}\textbf{100.0}  & 98.47  & \cellcolor{gray!10}98.78  & 94.18  & \cellcolor{gray!10}\underline{100.0}  & 64.52  & \cellcolor{gray!10}99.69   \\
        ~ & NMI & 20.60  & 21.42  & 91.72  & 100.0  & 89.41  & 29.53  & 31.23  & 53.35  & 25.79  & 43.37  & 15.87  & 98.73  & \cellcolor{gray!10}99.36 & 9.26  & \cellcolor{gray!10}\textbf{100.0}  & 97.04  & \cellcolor{gray!10}97.49  & 89.55  & \cellcolor{gray!10}\underline{100.0}  & 57.18  & \cellcolor{gray!10}99.37   \\
        ~ & ARI & 10.63  & 12.04  & 89.66  & 100.0  & 83.09  & 15.38  & 25.00  & 33.81  & 15.82  & 28.31  & 3.92  & 98.70  & \cellcolor{gray!10}99.33  & 2.85  & \cellcolor{gray!10}\textbf{100.0}  & 96.83  & \cellcolor{gray!10}97.48  & 87.50  & \cellcolor{gray!10}\underline{100.0}  & 45.28  & \cellcolor{gray!10}99.36   \\
        ~ & F1 & 31.27  & 31.00  & 92.63  & 100.0  & 82.64  & 31.39  & 24.41  & 45.62  & 33.25  & 47.05  & 22.68  & 99.34  & \cellcolor{gray!10}99.69  & 13.02  & \cellcolor{gray!10}\textbf{100.0}  & 98.35  & \cellcolor{gray!10}98.67  & 94.18  & \cellcolor{gray!10}\underline{100.0}  & 63.27  & \cellcolor{gray!10}99.66   \\
        \midrule[0.5pt] 
        \multirow{4}{*}{arXivAI} & ACC & 23.94  & 23.85  & 60.91  & 65.01  & 38.72  & \multirow{4}{*}{OOM} & \multirow{4}{*}{OOM} & 44.44  & 37.62  & \multirow{4}{*}{OOM} & 37.11  & 65.66  & \cellcolor{gray!10}\textbf{76.06}  & 30.71  & \cellcolor{gray!10}65.02  & 69.26  & \cellcolor{gray!10}\underline{70.58}  & 29.82  & \cellcolor{gray!10}58.75  & 28.45  & \cellcolor{gray!10}62.53   \\
        ~ & NMI & 1.10  & 10.20  & 34.34  & 36.18  & 32.54  & ~ & ~ & 21.63  & 20.73  & ~ & 2.24  & 39.24  & \cellcolor{gray!10}\textbf{45.29}  & 2.91  & \cellcolor{gray!10}36.19  & 36.18  & \cellcolor{gray!10}\underline{43.33}  & 1.28  & \cellcolor{gray!10}29.99  & 1.50  & \cellcolor{gray!10}33.07   \\
        ~ & ARI & 1.21  & 14.00  & 36.08  & 40.35  & 32.98  & ~ & ~ & 23.43  & 21.29  & ~ & 4.77  & 43.73  & \cellcolor{gray!10}\textbf{52.84}  & 5.35  & \cellcolor{gray!10}40.36  & 43.15  & \cellcolor{gray!10}\underline{49.48}  & 1.12  & \cellcolor{gray!10}32.97  & 1.51  & \cellcolor{gray!10}36.12   \\
        ~ & F1 & 19.25  & 19.20  & 49.47  & 53.66  & 16.97  & ~ & ~ & 33.96  & 31.62  & ~ & 20.94  & 52.86  & \cellcolor{gray!10}\textbf{60.06}  & 23.24  & \cellcolor{gray!10}52.66  & 54.55  & \cellcolor{gray!10}\underline{58.38}  & 19.22  & \cellcolor{gray!10}49.80  & 22.29  & \cellcolor{gray!10}50.66   \\
        \midrule[0.5pt] 
        \multirow{4}{*}{arXivCS} & ACC & 2.98  & 24.20  & \underline{34.42}  & 27.39  & \multirow{4}{*}{OOM} & \multirow{4}{*}{OOM} & \multirow{4}{*}{OOM} & 29.78  & 27.05  & \multirow{4}{*}{OOM} & \multirow{4}{*}{OOM} & 25.57  & \cellcolor{gray!10}\textbf{41.49}  & 11.27  & \cellcolor{gray!10}31.55  & 21.41  & \cellcolor{gray!10}29.45  & 8.94  & \cellcolor{gray!10}29.73  & 7.85  & \cellcolor{gray!10}30.08   \\
        ~ & NMI & 1.42  & 14.03  & 40.86  & 41.18  & ~ & ~ & ~ & 13.27  & 11.57  & ~ & ~ & 40.83  & \cellcolor{gray!10}\textbf{44.69}  & 5.12  & \cellcolor{gray!10}41.97  & 35.25  & \cellcolor{gray!10}\underline{43.20}  & 5.57  & \cellcolor{gray!10}39.98  & 3.84  & \cellcolor{gray!10}40.13   \\
        ~ & ARI & 1.32  & 11.80  & \underline{24.65}  & 19.14  & ~ & ~ & ~ & 14.32  & 12.02  & ~ & ~ & 16.51  & \cellcolor{gray!10}\textbf{35.04}  & 5.31  & \cellcolor{gray!10}22.58  & 14.59  & \cellcolor{gray!10}19.00  & 3.49  & \cellcolor{gray!10}18.86  & 2.48  & \cellcolor{gray!10}21.71   \\
        ~ & F1 & 2.20  & 12.33  & 20.39  & 21.41  & ~ & ~ & ~ & 14.08  & 13.28  & ~ & ~ & 19.56  & \cellcolor{gray!10}\textbf{27.31}  & 4.85  & \cellcolor{gray!10}\underline{23.15}  & 15.88  & \cellcolor{gray!10}22.68  & 4.02  & \cellcolor{gray!10}22.81  & 3.82  & \cellcolor{gray!10}22.69   \\
        \midrule[0.5pt] 
        \multirow{4}{*}{arXivMath} & ACC & 4.29  & 4.50  & 36.30  & 36.64  & \multirow{4}{*}{OOM} & \multirow{4}{*}{OOM} & \multirow{4}{*}{OOM} & \multirow{4}{*}{OOM} & \multirow{4}{*}{OOM} & \multirow{4}{*}{OOM} & \multirow{4}{*}{OOM} & 28.94  & \cellcolor{gray!10}\textbf{40.07}  & 5.82  & \cellcolor{gray!10}35.45  & 26.14  & \cellcolor{gray!10}\underline{39.74}  & 5.67  & \cellcolor{gray!10}29.93  & 5.43  & \cellcolor{gray!10}34.04   \\
        ~ & NMI & 1.44  & 1.43  & 35.38  & 40.42  & ~ & ~ & ~ & ~ & ~ & ~ & ~ & 37.32  & \cellcolor{gray!10}\textbf{42.52}  & 1.69  & \cellcolor{gray!10}38.97  & 26.16  & \cellcolor{gray!10}\underline{41.35}  & 1.39  & \cellcolor{gray!10}31.15  & 1.78  & \cellcolor{gray!10}39.03   \\
        ~ & ARI & 1.99  & 1.12  & 22.20  & 20.70  & ~ & ~ & ~ & ~ & ~ & ~ & ~ & 11.15  & \cellcolor{gray!10}\textbf{25.85}  & 1.38  & \cellcolor{gray!10}20.01  & 12.64  & \cellcolor{gray!10}\underline{22.68}  & 1.39  & \cellcolor{gray!10}16.11  & 1.22  & \cellcolor{gray!10}19.53   \\
        ~ & F1 & 3.51  & 3.59  & 28.94  & 31.53  & ~ & ~ & ~ & ~ & ~ & ~ & ~ & 26.87  & \cellcolor{gray!10}\textbf{33.91}  & 4.56  & \cellcolor{gray!10}31.10  & 22.90  & \cellcolor{gray!10}\underline{31.56}  & 4.43  & \cellcolor{gray!10}25.06  & 4.01  & \cellcolor{gray!10}29.27   \\ 
        \midrule[0.5pt] 
        \multirow{4}{*}{arXivPhy} & ACC & 2.61  & 2.38  & 31.39  & 31.84  & \multirow{4}{*}{OOM} & \multirow{4}{*}{OOM} & \multirow{4}{*}{OOM} & \multirow{4}{*}{OOM} & \multirow{4}{*}{OOM} & \multirow{4}{*}{OOM} & \multirow{4}{*}{OOM} & 30.01  & \cellcolor{gray!10}\underline{32.40}  & 6.63 & \cellcolor{gray!10}30.69 & 28.69  & \cellcolor{gray!10}30.37  & 5.60  & \cellcolor{gray!10}\textbf{32.96}  & 6.93  & \cellcolor{gray!10}30.64   \\ 
        ~ & NMI & 1.27  & 1.82  & 49.77  & 50.20  & ~ & ~ & ~ & ~ & ~ & ~ & ~ & \underline{50.40}  & \cellcolor{gray!10}\textbf{51.22}  & 3.82 & \cellcolor{gray!10}50.18 & 49.61  & 48.60  & 2.62  & \cellcolor{gray!10}35.65  & 2.53  & \cellcolor{gray!10}49.07   \\
        ~ & ARI & 1.40  & 1.63  & 21.78  & 21.46  & ~ & ~ & ~ & ~ & ~ & ~ & ~ & 20.15  & \cellcolor{gray!10}\textbf{22.86}  & 2.98& \cellcolor{gray!10}20.84 & 19.07  & \cellcolor{gray!10}20.00  & 1.10  & \cellcolor{gray!10}18.14  & 1.13  & \cellcolor{gray!10}\underline{22.15}   \\ 
        ~ & F1 & 1.82  & 1.99  & 23.92  & 23.58  & ~ & ~ & ~ & ~ & ~ & ~ & ~ & 22.45  & \cellcolor{gray!10}\underline{24.51}  & 4.43 & \cellcolor{gray!10}23.16 & 21.03  & \cellcolor{gray!10}22.30  & 2.94  & \cellcolor{gray!10}\textbf{28.90}  & 2.95  & \cellcolor{gray!10}23.19   \\
        \midrule[0.5pt] 
        \multirow{4}{*}{arXivLarge} & ACC & 1.13  & 1.82 & 22.72  & 22.28  & \multirow{4}{*}{OOM} & \multirow{4}{*}{OOM} & \multirow{4}{*}{OOM} & \multirow{4}{*}{OOM} & \multirow{4}{*}{OOM} & \multirow{4}{*}{OOM} & \multirow{4}{*}{OOM} & 21.18  & \cellcolor{gray!10}\textbf{26.97}  & 3.96 & \cellcolor{gray!10}\underline{22.77} & 7.00  & \cellcolor{gray!10}20.03  & 2.51  & \cellcolor{gray!10}20.60  & 4.46  & \cellcolor{gray!10}21.28   \\ 
        ~ & NMI & 1.17  & 1.30 & 50.16  & 51.39  & ~ & ~ & ~ & ~ & ~ & ~ & ~ & 50.84  & \cellcolor{gray!10}\textbf{52.89}  & 6.73 & \cellcolor{gray!10}\underline{51.54} & 9.68  & \cellcolor{gray!10}43.94  & 5.34  & \cellcolor{gray!10}48.55  & 4.23  & \cellcolor{gray!10}49.91   \\
        ~ & ARI & 1.50  & 1.48 & \underline{15.39} & 13.92  & ~ & ~ & ~ & ~ & ~ & ~ & ~ & 13.30  & \cellcolor{gray!10}\textbf{19.01}  & 3.43 & \cellcolor{gray!10}14.08 & 6.89  & \cellcolor{gray!10}13.80  & 1.60  & \cellcolor{gray!10}12.55  & 1.85  & \cellcolor{gray!10}12.29   \\
        ~ & F1 & 1.74  & 1.57 & 15.21  & 15.58  & ~ & ~ & ~ & ~ & ~ & ~ & ~ & 13.96  & \cellcolor{gray!10}\textbf{16.59}  & 2.68 & \cellcolor{gray!10}\underline{15.99} & 3.24  & 1\cellcolor{gray!10}0.99  & 1.22  & \cellcolor{gray!10}13.65  & 1.04  & \cellcolor{gray!10}15.69   \\
        \bottomrule[1.5pt]
    \end{tabular}}
\label{node clustering}
\end{table*}

\textbf{Q1:} \emph{Can BenchTGC Framework bring improvements?} \textbf{Answer:} \emph{Almost all temporal graph methods combined with BenchTGC Framework achieve excellent performance on all datasets.}

As shown in Table \ref{node clustering}, we report the node clustering performance of 21 methods on all 9 datasets:

(1) We mark the best reults in \textbf{bold} and the second best results in \underline{underline}. Thus there are 72 marks in the table, and we have 67 marks in our 5 improved methods, i.e., 93\% excellence rate. If we only focus on the best results, all red marks are in our improved methods, i.e., 100\% best rate. In addition, if BenchTGC brings improvements to the original methods, we gray out that results. Thus there are 180 comparisons, and only 2 comparisons showed no improvement, i.e., 99\% improvement rate. We achieve the best 42.15\% improvement on ``arXivCS'' dataset  (ARI metric from 24.65\% of Deepwalk to 35.04\% of IHTNE). Our improved IJODIE and ITREND methods achieve 100\% performance in all 4 metrics of School dataset.

(2) Obviously, all static graph clustering methods encounter OOM (Out-Of-Memory) problems on large-scale temporal graph datasets. Although different static methods have different ways of processing graph data, they are all train based on the adjacency matrix. This inevitably leads to the adjacency matrix scale expands exponentially as the node number increases. Although we can solve this problem by using CPU or multi-GPUs with larger memory, it still cannot cover up the limitations of static graph clustering: There must be a data scale that makes it impossible to train models based on the adjacency matrix, and this data scale is not uncommon in the information-based real world.

(3) When we focus on the improvements of BenchTGC Framework on temporal graph methods, we can find that some methods have decent results on common-scale datasets, but they do not perform well on large-scale datasets (whether the number of nodes or interactions is large), and their results are quite poor, such as JODIE, TREND, and S2T. This does not mean that the method design is bad, because these methods have been repeatedly verified. This just proves our point that the existing temporal graph methods mainly focus on link prediction and ignore node clustering. This leads to these methods being designed completely in a direction that is more suitable for link prediction. In addition, our BenchTGC framework can bring significant improvement by simply loading classic loss functions without precise design. 
Note that in these improved methods, not all modules are effective, which requires different model designs. For example, in JODIE, it proposes a unique t-batch technology to construct batches of different sizes, so the Cross-Batch Calibration module is difficult to work. In MNCI, since it focuses on community discovery, the module on cluster centers is redundant. However, there is still a lot of room for exploration on the combination of these modules, and it has not yet considered adjusting the weights of different modules to obtain better results. We have provided the specific model improvement details and module selection in the code \url{https://github.com/MGitHubL/BenchTGC}. Since this involves many implementation details, we will not introduce them one by one here.

(4) We would also want to point out that School dataset is the only special case in complexity level, namely the extremely rare case that we discussed above (i.e., $\mathcal{N}^2 < \mathcal{E}$). On this dataset, the static graph methods are significantly worse than temporal graph methods. Although JODIE is limited by the model design, with the improvements of our BenchTGC Framework, IJODIE and ITREND achieves 100\% significant performance on all 4 metrics. This supports our point that when the number of interactions exceeds the number of nodes, compressing the temporal graph data into static graph data will lose a large amount of effective information and bring about a significant decrease in performance. This further emphasizes the necessity and importance of temporal graph clustering.

\textbf{Q2}: \emph{Does BenchTGC Datasets make sense?} \textbf{Answer}: \emph{BenchTGC Datasets reflect the OOM problem of static graph clustering, and has reliable labels than public large-scale datasets.}

(5) Static graph clustering methods such as MVGRL, SDCN, and DFCN perform well on common-scale datasets, and some of the results are excellent. However, this cannot cover up a defect of static graph clustering, that is, the training pattern based on the adjacency matrix will encounter the problem of excessive memory. The BenchTGC Datasets we developed introduce several datasets that are larger in scale both in terms of node number and edge number, which resulted in the phenomenon that all static graph clustering methods eventually encountered OOM problem.

(6) In addition, other methods have good clustering performance on these datasets, so that we can judge whether a model is really effective from the changes in experimental results. This conclusion is drawn through the comparison of datasets. Note that in Table \ref{meta}, we report the results on Meta dataset. Such dataset can represent those we classified as "clustering inapplicable datasets". On Meta, we select three different sets of category features as labels, and executed the classic method (node2vec), static graph clustering method (SDCN), and temporal graph learning method (HTNE), respectively. It can be seen that the results of NMI and ARI are very poor. This situation is avoided on the "clustering use dataset", i.e., the datasets we used in Table \ref{node clustering}. Although the ARI results of some methods are also relatively low on some large-scale datasets, compared with the almost invisible values on Meta, they can still be distinguished by comparing with the performance of other methods. Thus we argue that the advantage of these datasets is that they can obtain relatively reliable node labels, so that the evaluation metrics can fluctuate normally (rather than all methods are very low as on Meta). Therefore, we believe the developed datasets are labeled reliably.

\subsection{GPU Memory Analysis}

\begin{figure}[t]
    \centering
    \includegraphics[width=0.48\textwidth]{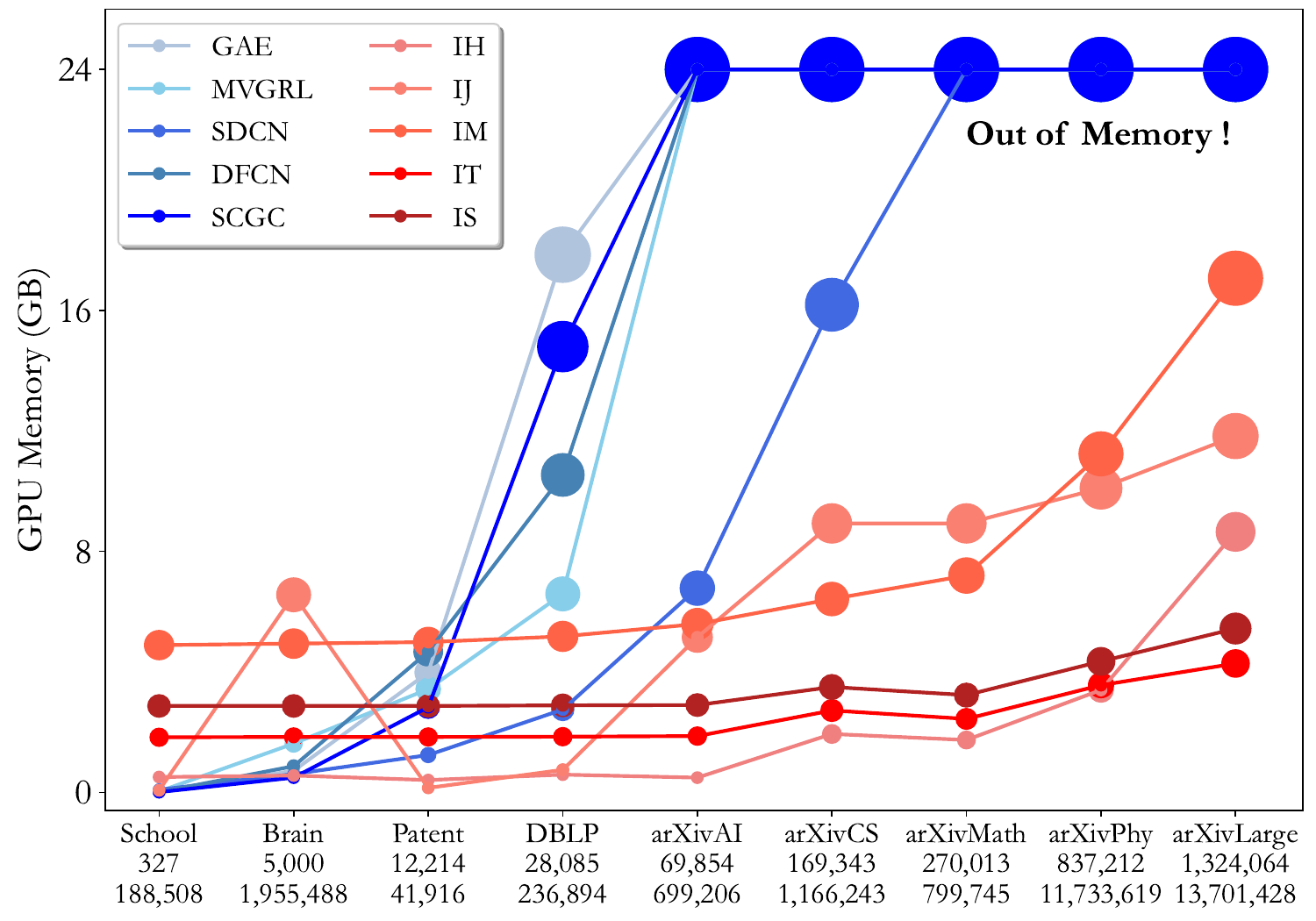}
    \caption{GPU memory usage of static methods and temporal methods. We give the number of nodes and interactions for each dataset, and these datasets are sorted by node numbers. When a method encounters OOM problem, it will report an error. We mark this state as 24 GB, which is the upper bound of our GPU.}
    \label{allGPU}
\end{figure}

\textbf{Q3:} \emph{How does TGC reduce memory pressure?} \textbf{Answer:} \emph{TGC can maintain a small memory on large-scale data with the batch-processing pattern based on interaction sequence.}

In order to explore the superiority of TGC in terms of memory usage, we select static methods (i.e., GAE, MVGRL, SDCN, DFCN, and SCGC) and temporal methods (i.e., IHTNE, IJODIE, IMNCI, ITREND, and IS2T) for comparison. Among them, we use the improved versions of temporal graph methods, because these versions added new loss functions and often required larger memory. In Figure \ref{allGPU}, We report the GPU memory usage between static graph methods and temporal graph methods. We can find that:

(1) All static graph clustering methods have OOM problems when encountering large-scale datasets, but this phenomenon has never occurred in temporal graph clustering methods. It should be noted that for all temporal graph methods, we used the batch size of 10,000. In fact, this size is not common in actual training, and is often 128 or 1024. A larger batch size will lead to a larger memory usage, but even in this case, the memory requirements of temporal methods are still small. This means that TGC has a natural advantage in processing large-scale data. (The effect of batch size on memory and time will be discussed in next subsection.)

(2) Compared with Table \ref{node clustering}, we can find that although some methods exceed the GPU memory, we still report their experimental results. These cases are trained on the CPU, which will lead to a very slow training speed, thus losing the powerful support of GPU for deep learning. This means that using the CPU for training is not a good way to solve the problem, and the dataset scale that can use the CPU is also limited. When the data scale is further increase, the OOM problem will still occur, which is inevitable.

(3) This result also echoes what we pointed out above, in most cases, $\mathcal{N}^2 > \mathcal{E}$. When a temporal graph method is trained in batches, obviously, $\mathcal{N}^2 \gg \mathcal{B}$. For example, in arXivCS dataset, all static methods encounter OOM problem except SDCN with 16.19 GB. In contrast, the maximum memory usage in temporal methods is only 8.93 GB (IJODIE), and the minimum memory usage is only 1.94 GB (IHTNE).

(4) Since K-means and AE do not use graph structures, and Deepwalk and node2vec are trained using CPU, we do not compare them here. Note that these methods can also run well on large-scale datasets, and some have shown excellent results, so they can also serve as one of the breakthrough points for future research.

\subsection{Time-Space Balance Analysis}

\begin{figure*}[t]
    \centering
    \begin{minipage}[t]{0.25\textwidth}
        \includegraphics[width=1\textwidth]{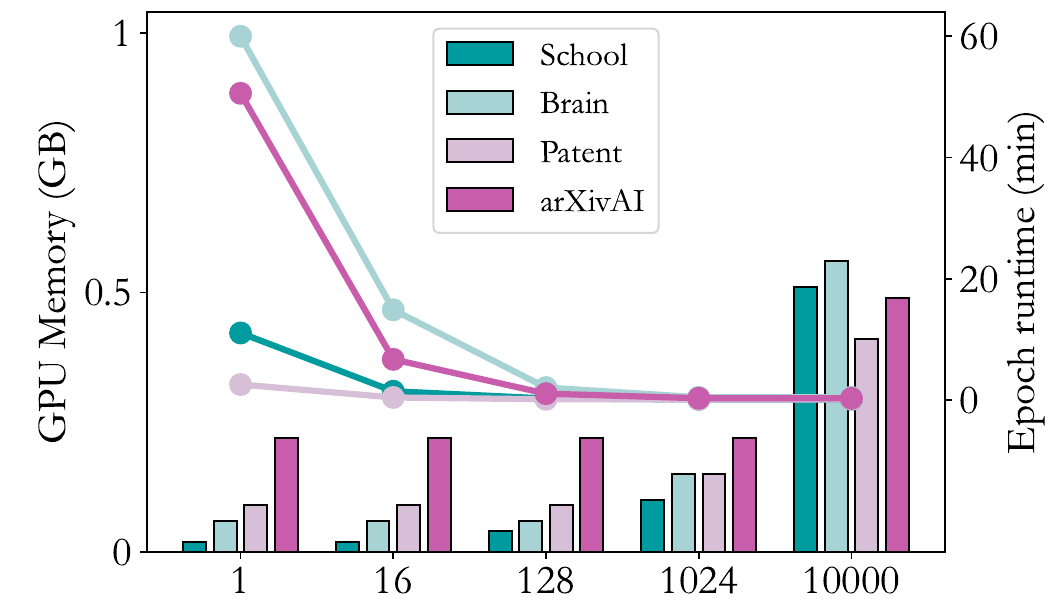}
        \centerline{(a) IHTNE}
    \end{minipage}%
    \begin{minipage}[t]{0.25\textwidth}
        \includegraphics[width=1\textwidth]{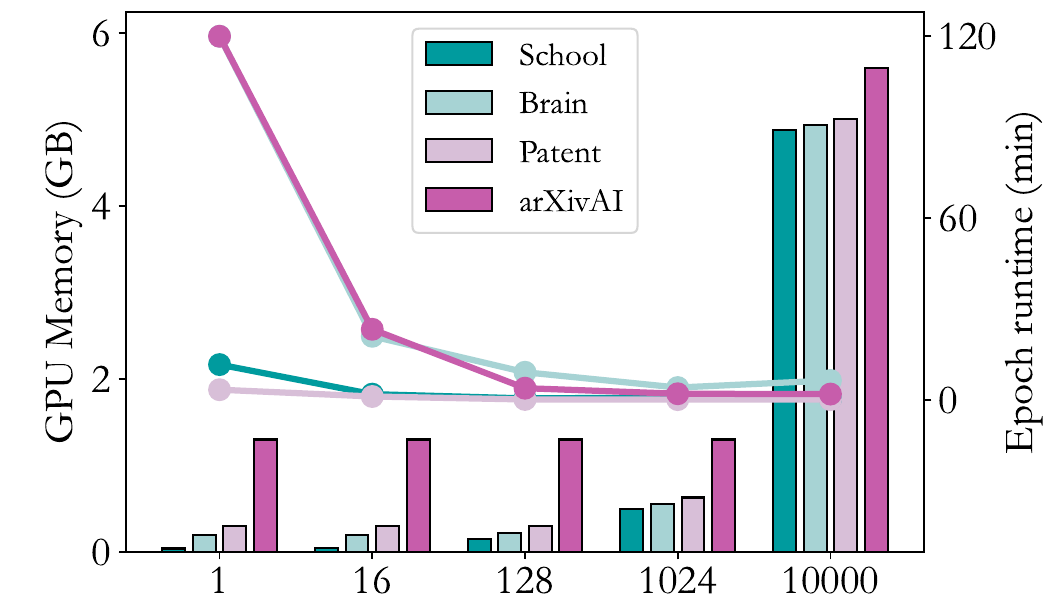}
        \centerline{(b) IMNCI}
    \end{minipage}%
    \begin{minipage}[t]{0.25\textwidth}
        \includegraphics[width=1\textwidth]{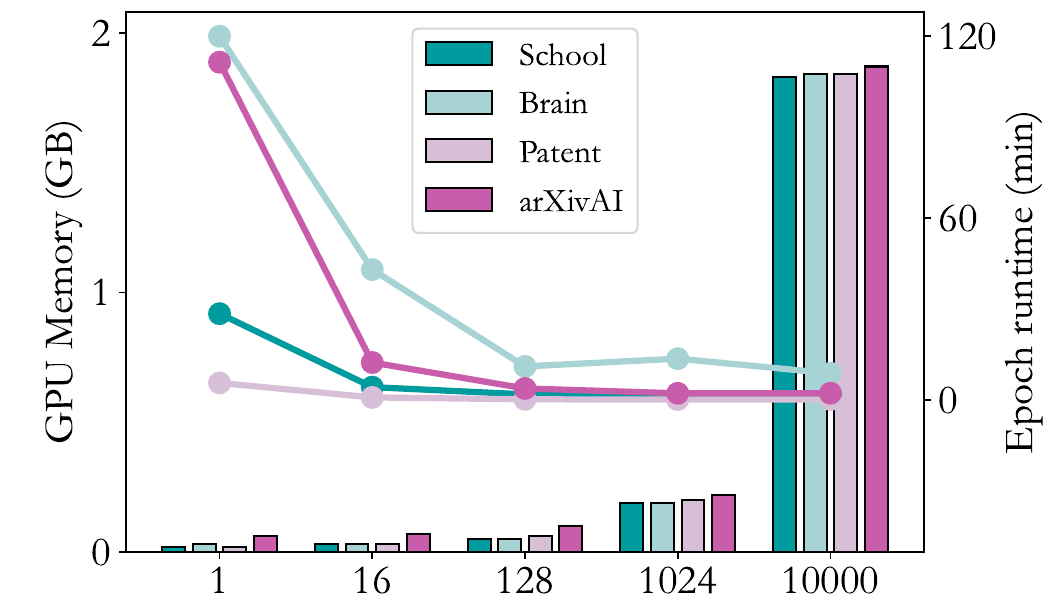}
        \centerline{(c) ITREND}
    \end{minipage}%
    \begin{minipage}[t]{0.25\textwidth}
        \includegraphics[width=1\textwidth]{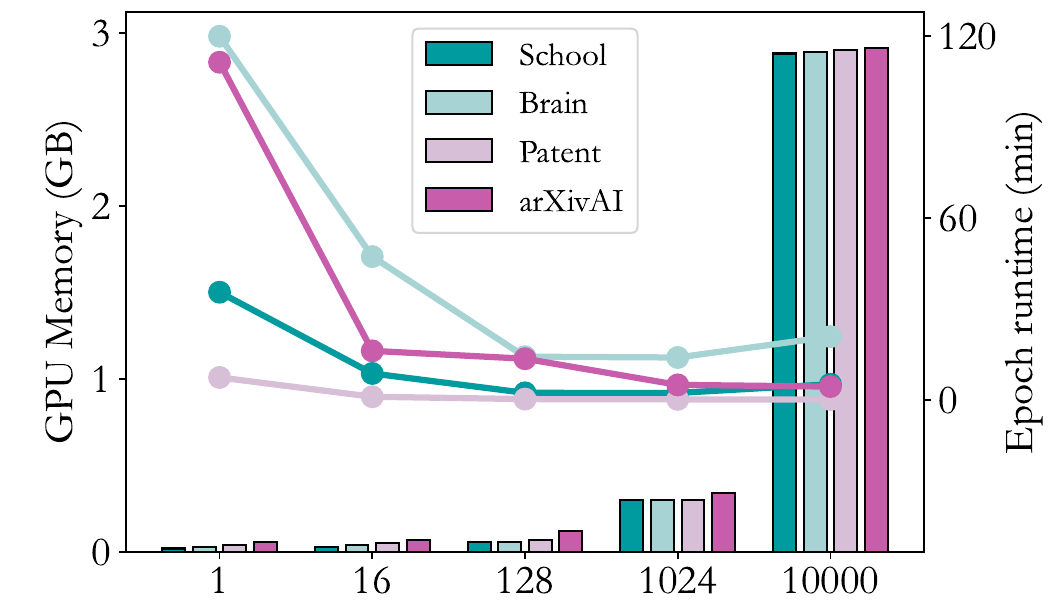}
        \centerline{(d) IS2T}
    \end{minipage}%
    \caption{Runtime and GPU memory changes in one epoch with different batch sizes. The bars denote GPU usage, and the lines denote runtime.}
    \label{t and s}
\end{figure*}

\textbf{Q4:} \emph{How is the flexibility of TGC reflected?} \textbf{Answer:} \emph{TGC can find the balance between time and space requirements by adjusting the batch size in training, i.e., Time-Space Balance.}

To further discuss the flexibility of TGC, we conduct experiments to observe the runtime and GPU memory changes in one epoch with different batch sizes. Here we observe 4 improved temporal graph methods except IJODIE. Due to JODIE utilizes a ``t-batch'' module to adjust batch size according to datasets adaptively, we can not adjust its batch size. As shown in Figure \ref{t and s}, we set batch sizes as 1, 16, 128, 1024, and 10000, and select 4 datasets (School, Brain, Patent, and arXivAI) for experiments. We can find that:

(1) When batch size is very small (especially as 1), all of the GPU memory usages are smaller than 0.5 GB. It means that even the largest datasets (even ones 10 to 100 times larger than arXivLarge) can be run on almost any GPU. The batch size achieves 1 means only one interaction in each batch training, and it will consume a lot of time. All 4 improved methods encountered timeout issues on the brain dataset. Among them, IHTNE exceeds 60 minutes, and the other 3 methods exceed 120 minutes, which we mark as reaching the upper limit in the figure (i.e., Out of Memory!).

(2) When the batch size becomes large (from 128 to 10000), the GPU memory usuage starts to grow and suddenly becomes more larger at 10000 (when batch size as 100000, OOM problem occurs). Meanwhile, the runtime of one epoch begins to stabilize and less than 5 minutes.

(3) This means that temporal graph clustering methods have strong flexibility in different scenarios. Since the memory usage and runtime can be changed by adjusting the batch size, a dynamic balance can be found so that methods can meet the GPU requirements without taking too long to run. In other words, we argue that \textbf{temporal graph clustering can find a balance between time and space requirements}.

(4) This also echoes the above GPU analysis. As can be seen from the Figure \ref{t and s}, when the batch size reaches 10000, the amount of GPU memory consumed is far beyond the commonly used sizes of 128 and 1024. Even in this case, the temporal graph clustering methods can still maintain a relatively healthy memory usage when facing large-scale datasets, compared with the OOM problem of static graph clustering. So when the batch size is adjusted to 1024 in actual training, datasets that are hundreds of times larger than arXivLarge dataset can still be trained smoothly.

\subsection{Ablation Study}

\textbf{Q5:} \emph{How to use the BenchTGC Framework?} \textbf{Answer:} \emph{For different data distributions, the pre-trained features can be considered combining with different loss functions.
}

As mentioned above, the temporal graph clustering process contains different 3 stages: pre-processing, training, and clustering. Among them, since one-step clustering is still a preliminary concept, we mainly explore ablation experiments on pre-processing and training. 

\subsubsection{Pre-Processing Feature Selection}

\begin{table}[t]
\centering
\caption{Ablation study on initial features. ``H+G'' denotes HTNE training on generated feature, ``T+P'' denotes TREND training on pre-trained feature. We labeled the better result compared to both as gray.}
\label{tab:ab feature}

\begin{tabular}{cc|cc|cc}
\toprule[1.5pt]
Data& Metric &H+G  &H+P  &T+G  &T+P \\
\midrule[0.5pt]
\multirow{4}{*}{DBLP}&  ACC&  45.74&  \cellcolor{gray!10}48.32&  46.82& \cellcolor{gray!10}49.00\\
~&  NMI&  35.95&  \cellcolor{gray!10}36.75&  36.56& \cellcolor{gray!10}39.24\\
~&  ARI&  22.13&  \cellcolor{gray!10}22.31&  22.83& \cellcolor{gray!10}26.67\\
~&  F1&  43.98&  \cellcolor{gray!10}44.48&  44.54& \cellcolor{gray!10}44.82\\
\midrule[0.5pt]
\multirow{4}{*}{Brain}&  ACC&  \cellcolor{gray!10}43.20& 43.18 &  39.83& \cellcolor{gray!10}40.14\\
~&  NMI&  50.33&  \cellcolor{gray!10}50.80&  \cellcolor{gray!10}45.64& 43.41\\
~&  ARI&  29.26&  \cellcolor{gray!10}29.95&  22.82& \cellcolor{gray!10}24.52\\
~&  F1&  \cellcolor{gray!10}43.85&  42.28&  33.67& \cellcolor{gray!10}40.40\\
\midrule[0.5pt]
\multirow{4}{*}{School}&  ACC&  99.38&  \cellcolor{gray!10}99.39&  94.18& \cellcolor{gray!10}97.85\\
~&  NMI&  98.73&  \cellcolor{gray!10}98.74&  89.55& \cellcolor{gray!10}96.06\\
~&  ARI&  98.70&  \cellcolor{gray!10}98.71&  87.50& \cellcolor{gray!10}95.60\\
~&  F1&  99.34& \cellcolor{gray!10}99.35&  94.18& \cellcolor{gray!10}97.65\\
\midrule[0.5pt]
\multirow{4}{*}{arXivCS}&  ACC&  25.57&  \cellcolor{gray!10}37.86&  8.94& \cellcolor{gray!10}16.01\\
~&  NMI&  40.83&  \cellcolor{gray!10}44.51&  5.57& \cellcolor{gray!10}25.88\\
~&  ARI&  16.51&  \cellcolor{gray!10}30.34&  3.49& \cellcolor{gray!10}9.62\\
~&  F1&  19.56&  \cellcolor{gray!10}26.41&  4.02& \cellcolor{gray!10}10.07\\
\bottomrule[1.5pt]
\end{tabular}
\end{table}

\begin{figure*}[t]
    \centering
    \begin{minipage}[t]{0.45\textwidth}
    \includegraphics[width=1\textwidth]{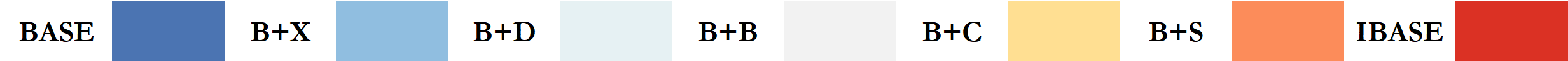}
    \end{minipage}%
    \vspace{1.5mm}
    \hfill
    \begin{minipage}[t]{0.25\textwidth}
        \includegraphics[width=1\textwidth]{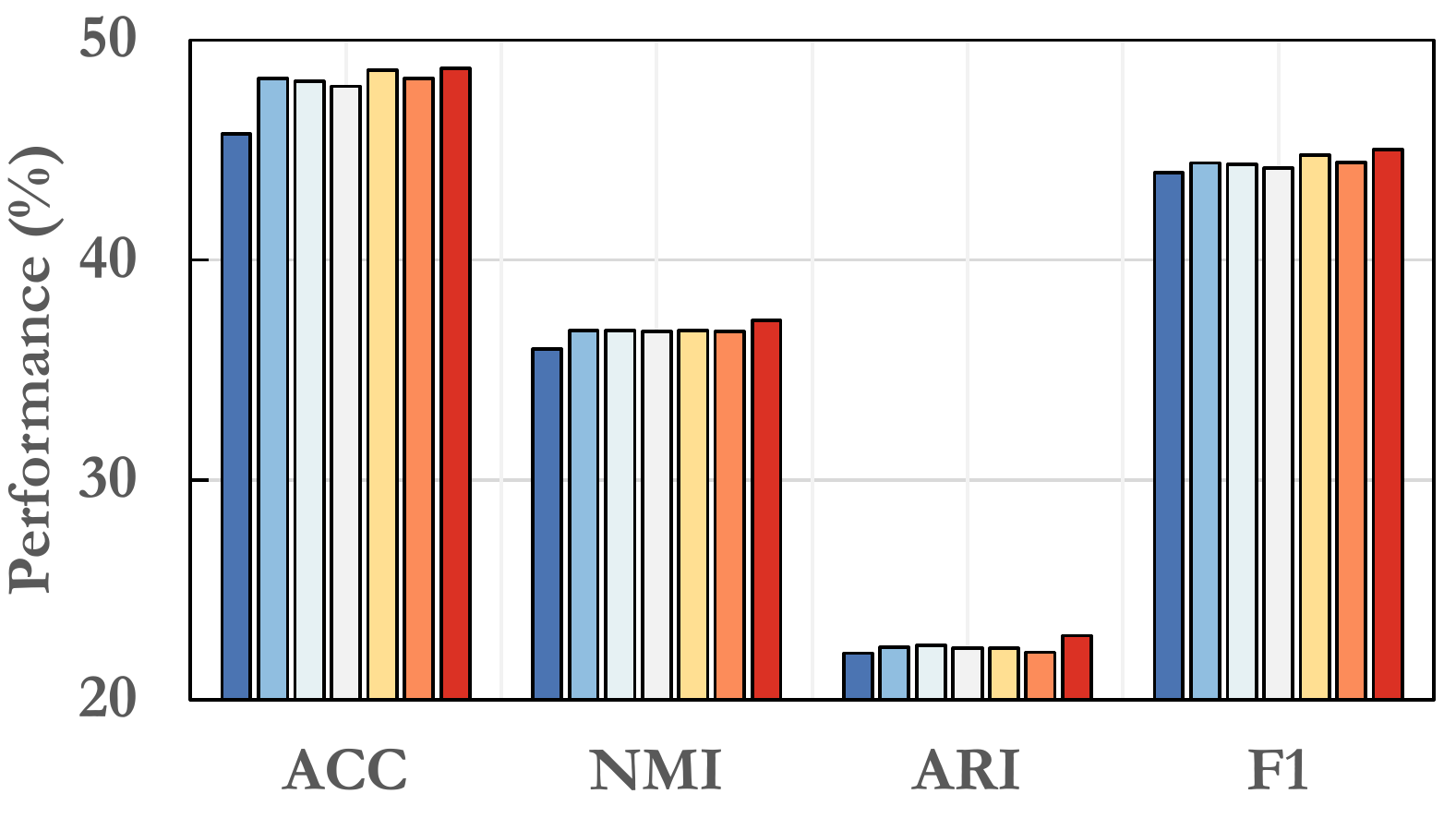}
        \centerline{\quad(a) HTNE with DBLP}
        \newline
    \end{minipage}%
    \begin{minipage}[t]{0.25\textwidth}
        \includegraphics[width=1\textwidth]{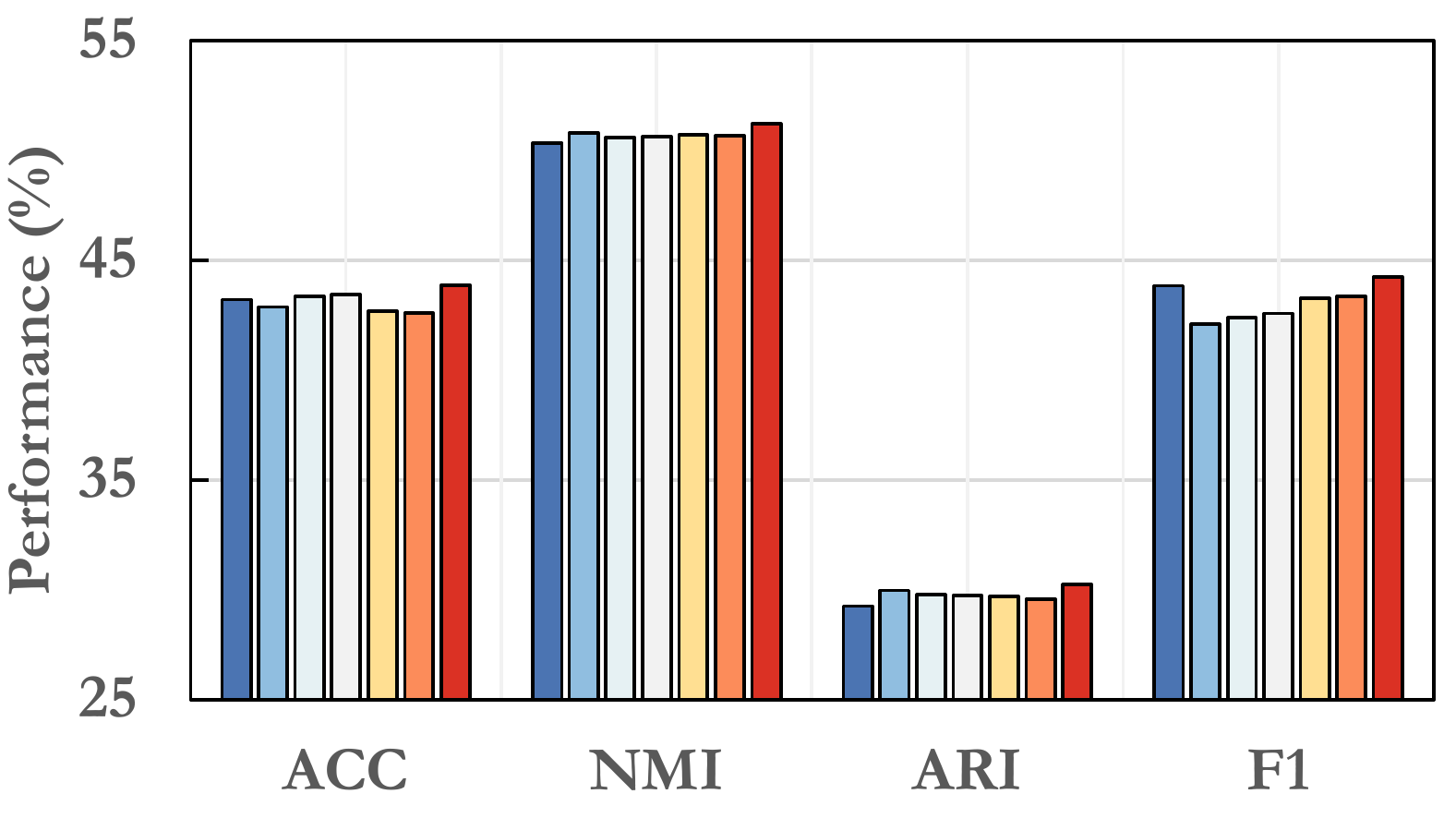}
        \centerline{\quad(b) HTNE with Brain}
        \newline
    \end{minipage}%
    \begin{minipage}[t]{0.25\textwidth}
        \includegraphics[width=1\textwidth]{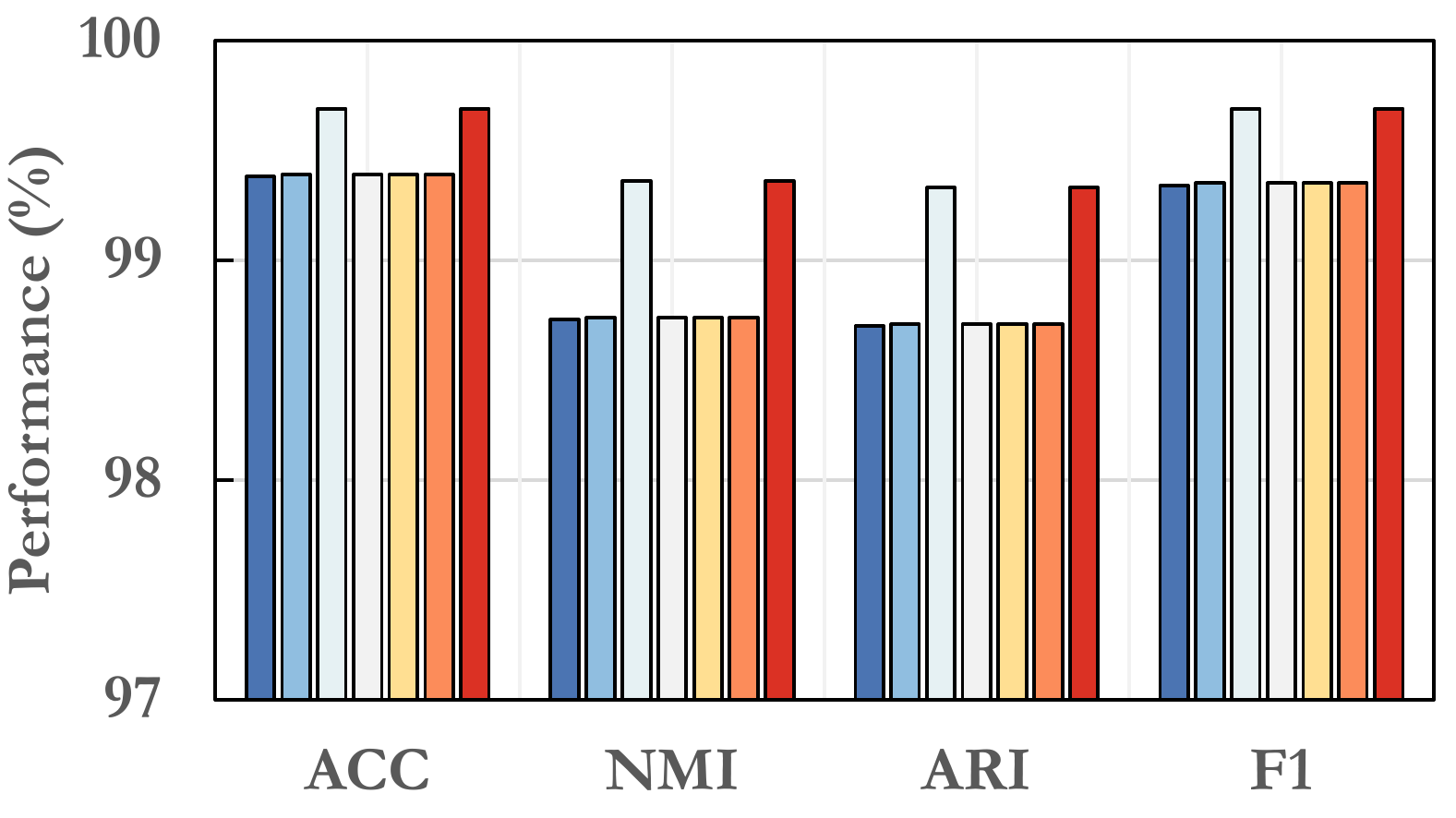}
        \centerline{\quad(c) HTNE with School}
        \newline
    \end{minipage}%
    \begin{minipage}[t]{0.25\textwidth}
        \includegraphics[width=1\textwidth]{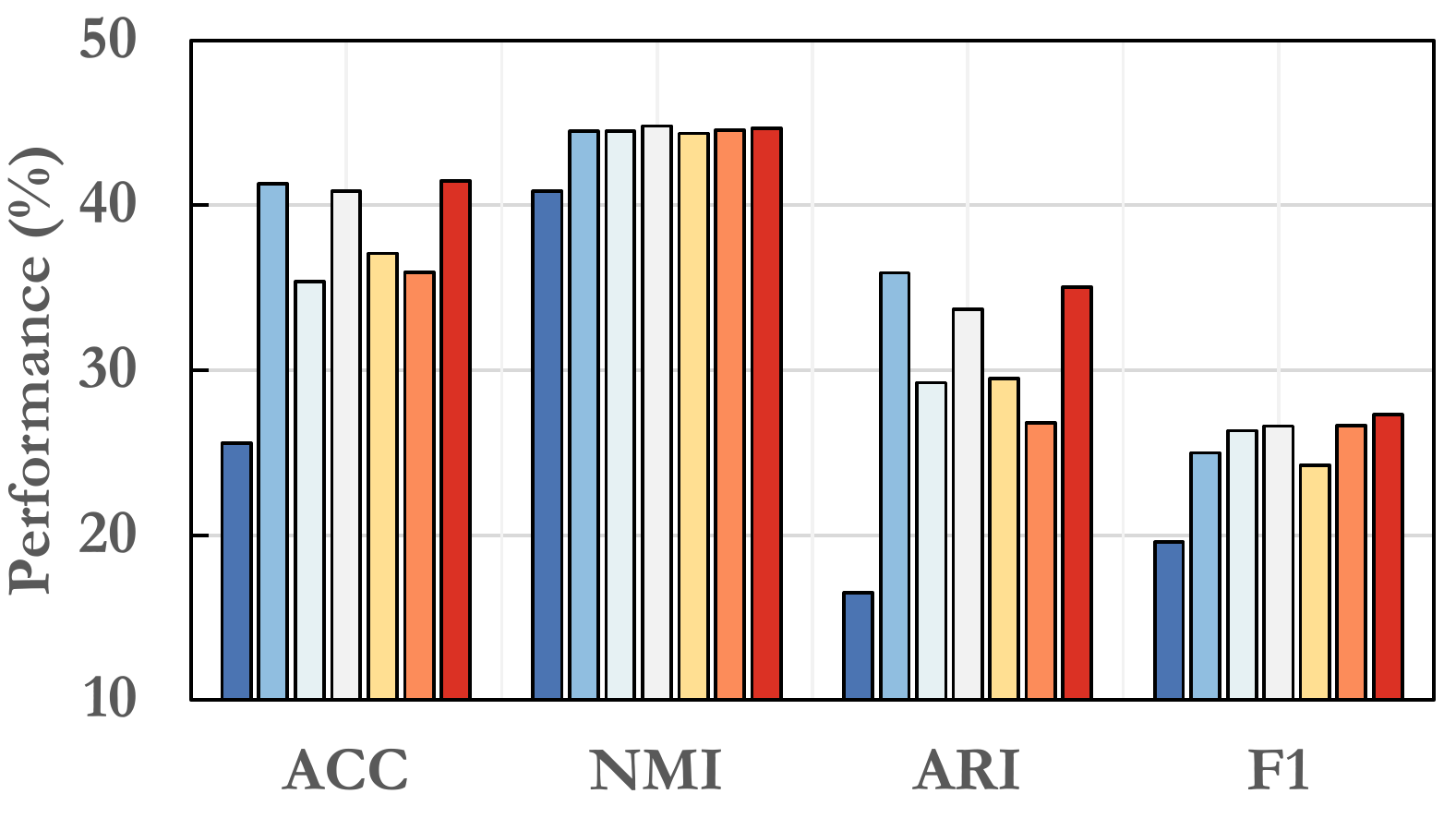}
        \centerline{\quad(d) HTNE with arXivCS}
        \newline
    \end{minipage}%
    \newline
    \begin{minipage}[t]{0.25\textwidth}
        \includegraphics[width=1\textwidth]{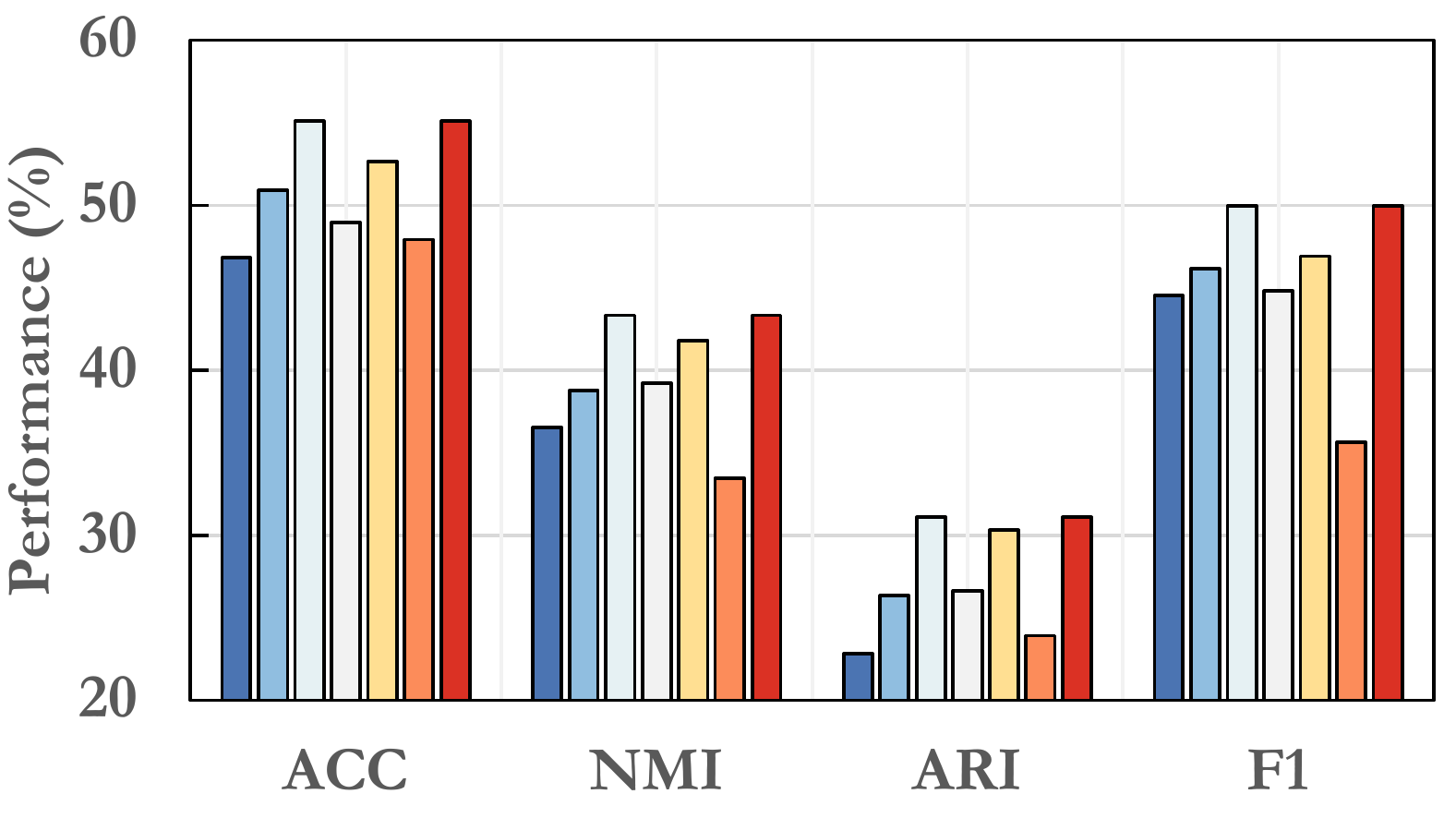}
        \centerline{\quad(e) TREND with DBLP}
    \end{minipage}%
    \begin{minipage}[t]{0.25\textwidth}
        \includegraphics[width=1\textwidth]{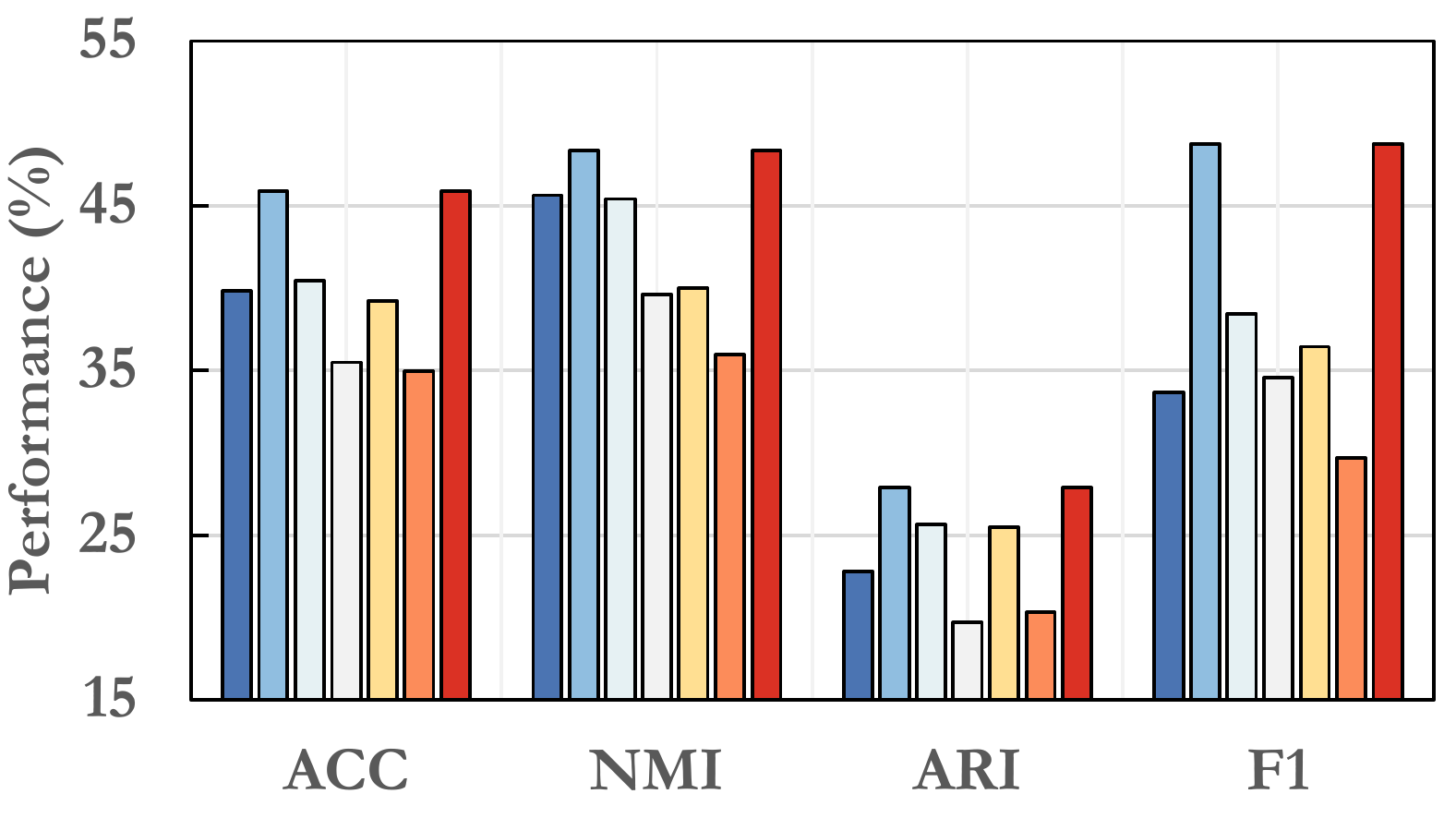}
        \centerline{\quad(f) TREND with Brain}
    \end{minipage}%
    \begin{minipage}[t]{0.25\textwidth}
        \includegraphics[width=1\textwidth]{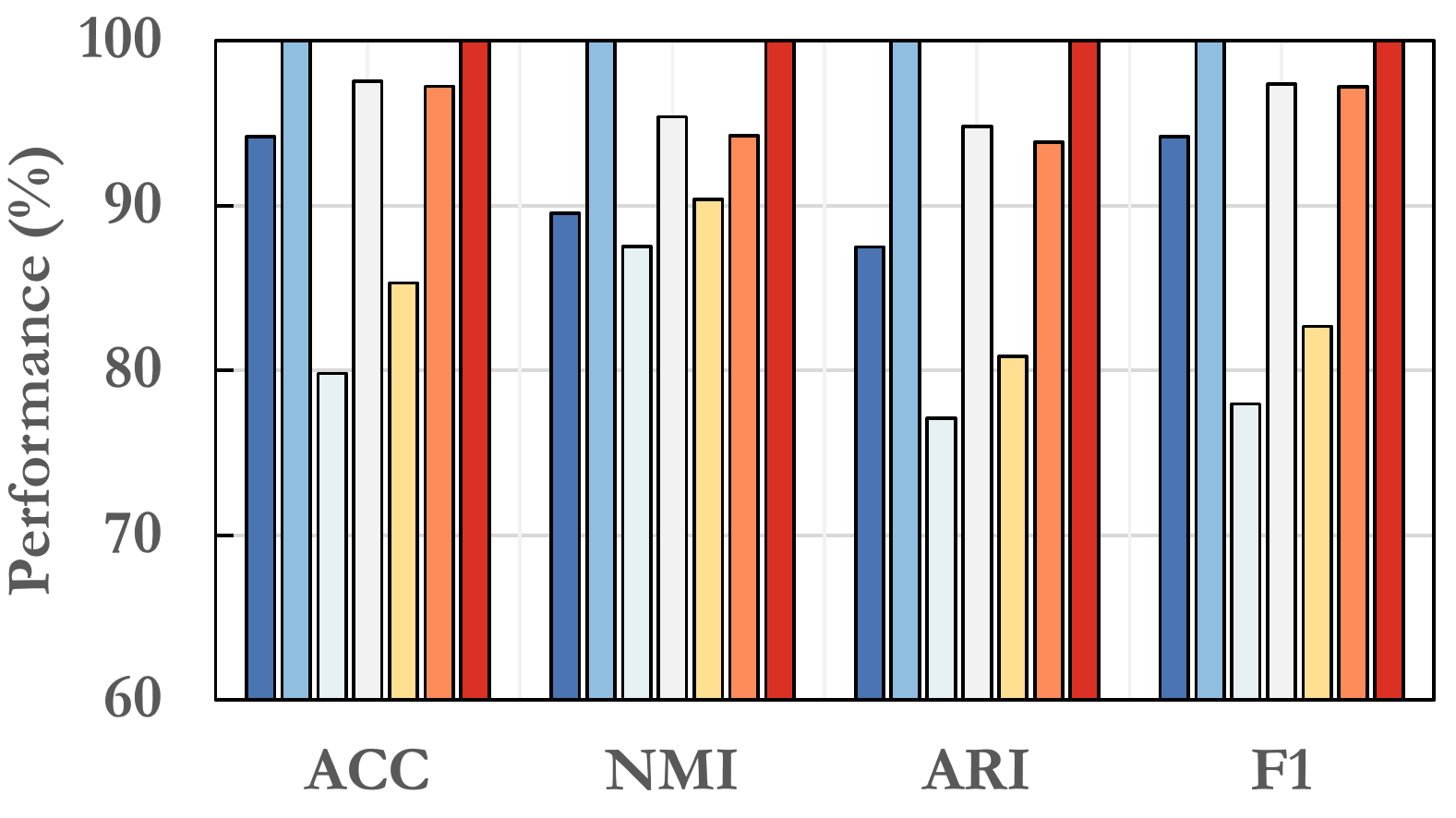}
        \centerline{\quad(g) TREND with School}
    \end{minipage}%
    \begin{minipage}[t]{0.25\textwidth}   \includegraphics[width=1\textwidth]{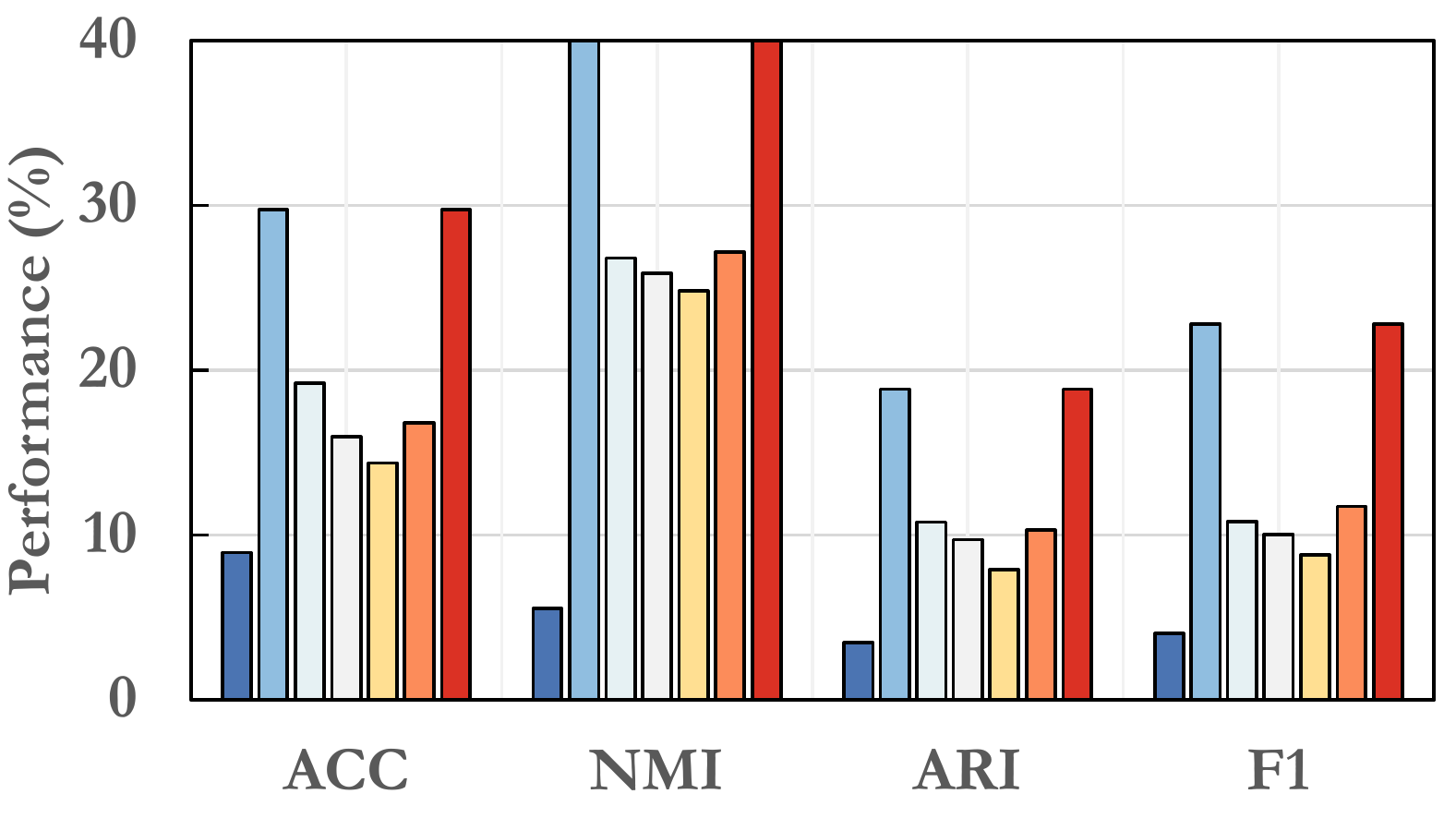}
        \centerline{\quad(h) TREND with arXivCS}
    \end{minipage}%
    \caption{Ablation study on different loss functions. For simplicity, we denote HTNE and TREND as ``BASE''. For different loss functions, we use their subscript abbreviations to refer to them, such as ``$\mathcal{L}_X$'' as X.}
    \label{ab loss}
\end{figure*}

Initial feature is important which can help methods avoid cold-starting problem, but in many public temporal graph datasets, there are no initial features. The datasets used in our experiments also meet this problem, all of 9 datasets have no features to load. We have discussed the difference between feature generation and pre-training in the Framework, and here we will discuss the advantages of pre-trained features over generated features on clustering performance.

As shown in Table \ref{tab:ab feature}, we report the performance of HTNE and TREND on 4 datasets. To save space, we denote ``H'' as HTNE, ``T'' as TREND, ``G'' as generated features, and ``P'' as pre-trained features. We labeled the better result compared to both as gray. To ensure the comprehensiveness of the comparison, we used random initialization features on the HTNE method and positional encoding features on TREND. The one-hot embedding is usually not considered even though it gives better results because it takes up too much memory ($\mathbb{R}^{\mathcal{N} \times \mathcal{N}}$) on large-scale datasets.

Through the results, we can find that almost in all cases, pre-trained features can bring better performance for methods that generated features. This of course also depends on the effectiveness of the pre-trained model, but in general, the node embeddings obtained after a round of training tend to have richer and more reliable information than the initialized features, especially when those classical methods are used as pre-trained modules. Note that pre-training is based on the premise that the original features are unavailable or unreliable. When such features are meaningful, pre-training is just an option. Compared with generated features, pre-trained features achieve better results because they have undergone autonomous learning of the classic model. To some extent, this is similar to the concept of knowledge distillation. Through the transformation or denoising of the initial data by the classic model, more information-rich features can be extracted, which is the significance of pre-training. Under this premise, researchers are very free to select what technology to use in pre-training.

\subsubsection{Training Loss Selection}

Here we discuss the effects of different loss functions proposed in BenchTGC Framework on the clustering performance. Figure \ref{ab loss} reports the results of HTNE and TREND on 4 datasets (DBLP, Brain, School, and arXivCS). We can find that these different loss functions are effective on different indicators and datasets. Specifically, $\mathcal{L}_X$ and $\mathcal{L}_D$ are often effective, while other loss functions may have their own suitable data distribution. However, this statement is not absolute, because better benefits can be obtained by combining different loss functions. For example, IHTNE achieved best results on DBLP and Brain datasets by combining all these loss functions together.


Furthermore, combined with the data distribution, we discuss the selection of different modules. Taking DBLP as an example, this type of academic datasets are the basis of most graph methods, that is, the appearance of interactions is often presented in the form of small circles. In this distribution, there are naturally some central nodes that gather information from others, so distribution alignment loss is not so important. Instead, the cluster scaling loss can clearly demarcate their boundaries. In contrast, the School dataset is a frequent interaction between students. Even if there are some central nodes, they are not very prominent because the number of interactions of each node is too large. In this case, the distribution alignment loss can better help the model find the right clustering position. Therefore, this echoes what we have repeatedly mentioned that the choice of clustering module depends on the data distribution.





\subsection{Visualization}

Visualization is a classic experiment in clustering task, and here we visualize the clusters and distribution.

\subsubsection{Clusters visualization}

\begin{figure*}[t]
    \centering
    \begin{minipage}[t]{0.125\textwidth}
        \includegraphics[width=1\textwidth]{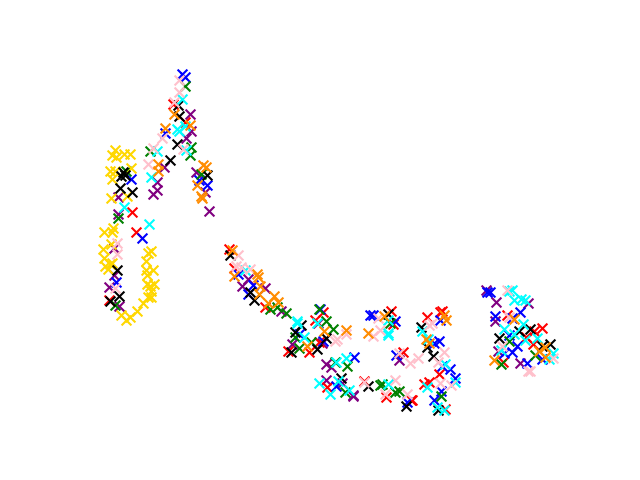}
        \newline
    \end{minipage}%
    \begin{minipage}[t]{0.125\textwidth}
        \includegraphics[width=1\textwidth]{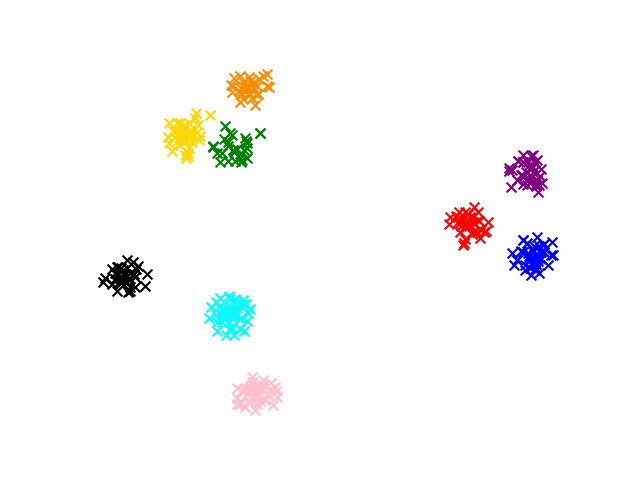}
        \newline
    \end{minipage}%
    \begin{minipage}[t]{0.125\textwidth}
        \includegraphics[width=1\textwidth]{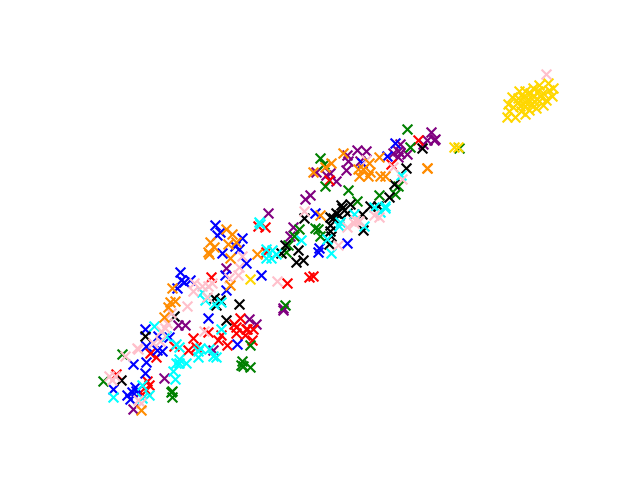}
        \newline
    \end{minipage}%
    \begin{minipage}[t]{0.125\textwidth}
        \includegraphics[width=1\textwidth]{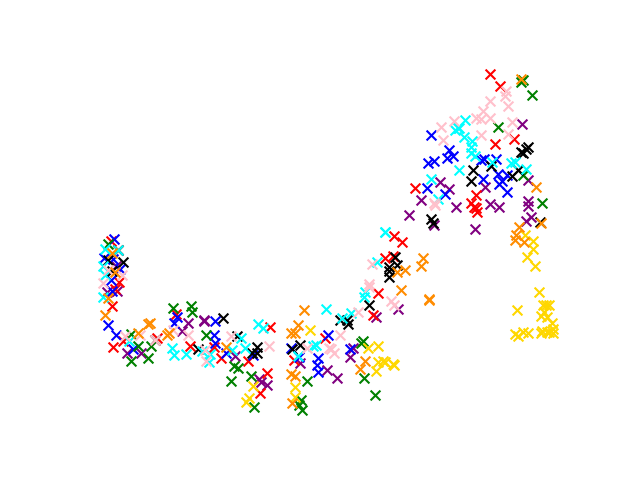}
        \newline
    \end{minipage}%
    \begin{minipage}[t]{0.125\textwidth}
        \includegraphics[width=1\textwidth]{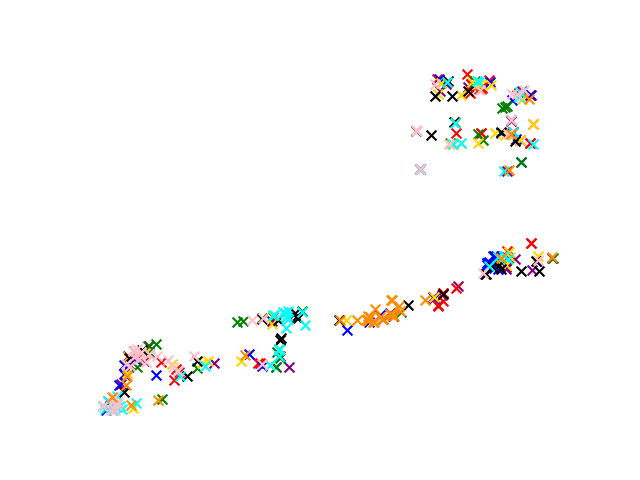}
        \newline
    \end{minipage}%
        \begin{minipage}[t]{0.125\textwidth}
        \includegraphics[width=1\textwidth]{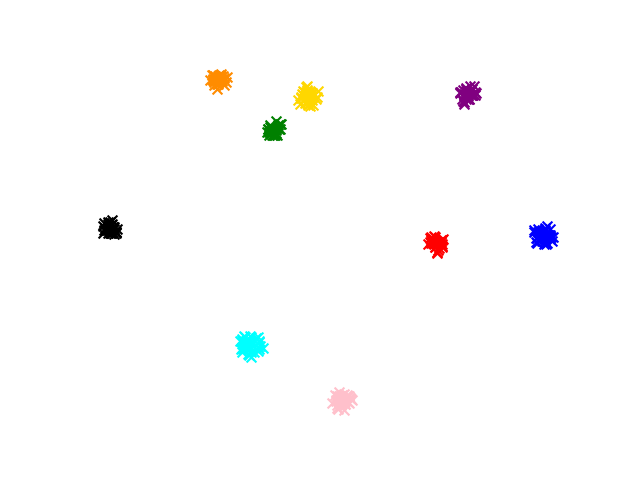}
        \newline
    \end{minipage}%
    \begin{minipage}[t]{0.125\textwidth}
        \includegraphics[width=1\textwidth]{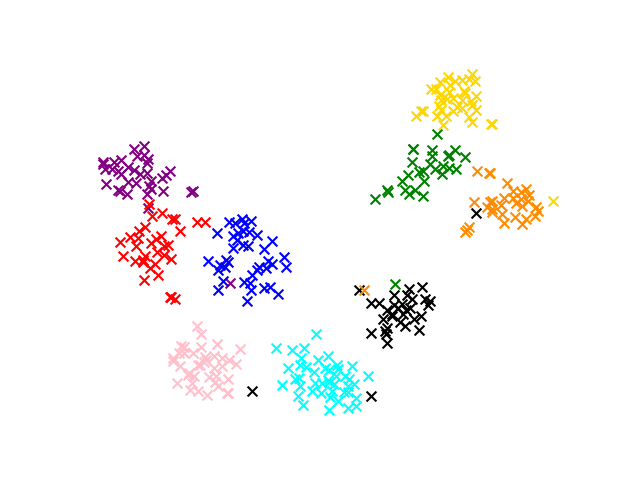}
        \newline
    \end{minipage}%
    \begin{minipage}[t]{0.125\textwidth}
        \includegraphics[width=1\textwidth]{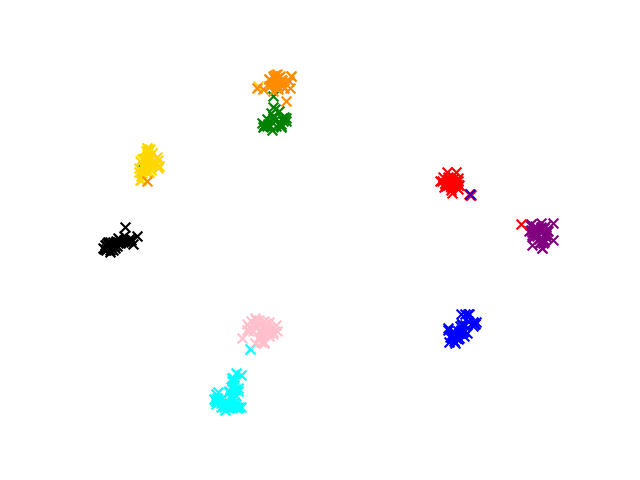}
        \newline
    \end{minipage}%
    \newline
    \begin{minipage}[t]{0.125\textwidth}
        \includegraphics[width=1\textwidth]{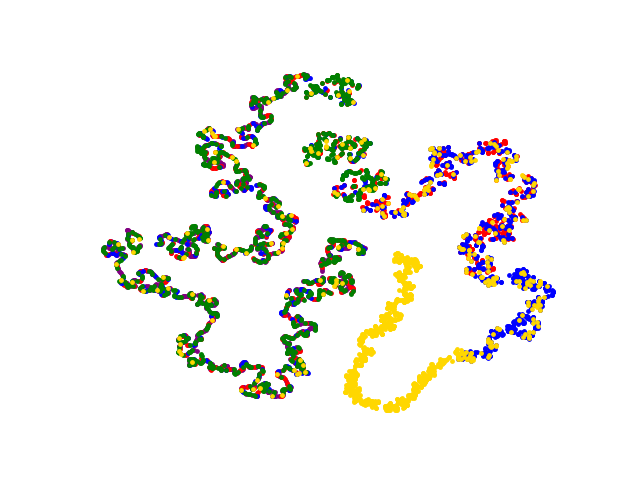}
        \centerline{(a) AE}
    \end{minipage}%
    \begin{minipage}[t]{0.125\textwidth}
        \includegraphics[width=1\textwidth]{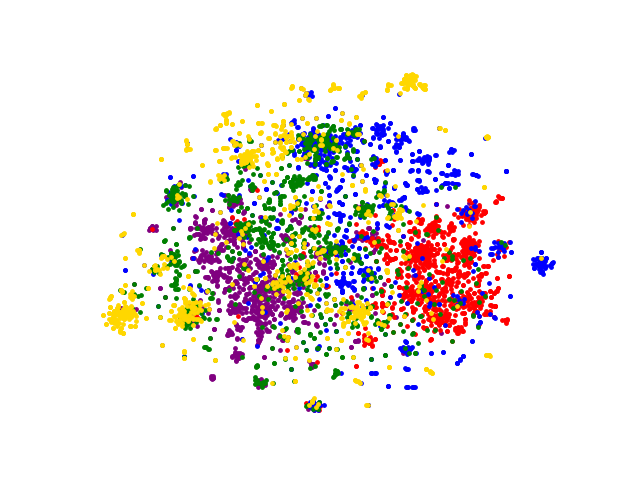}
        \centerline{(b) DW}
    \end{minipage}%
    \begin{minipage}[t]{0.125\textwidth}
        \includegraphics[width=1\textwidth]{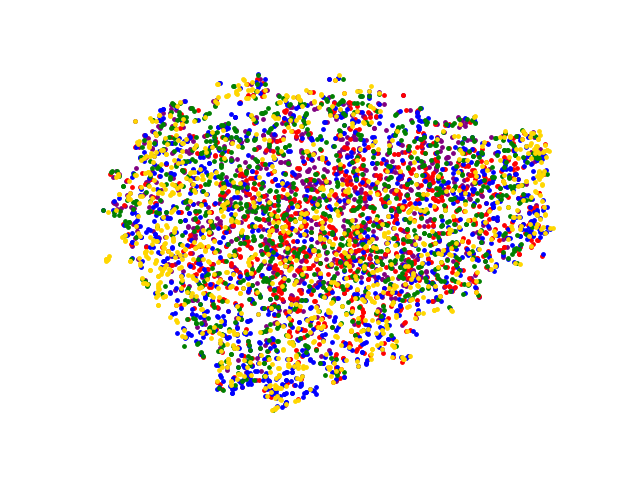}
        \centerline{(c) SDCN}
    \end{minipage}%
    \begin{minipage}[t]{0.125\textwidth}
        \includegraphics[width=1\textwidth]{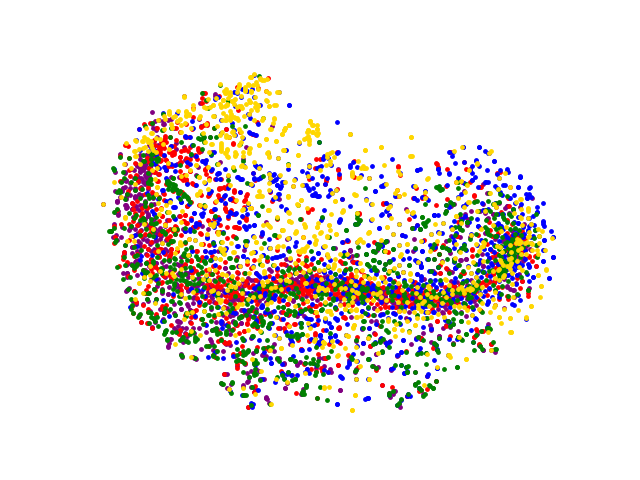}
        \centerline{(d) SCGC}
    \end{minipage}%
    \begin{minipage}[t]{0.125\textwidth}
        \includegraphics[width=1\textwidth]{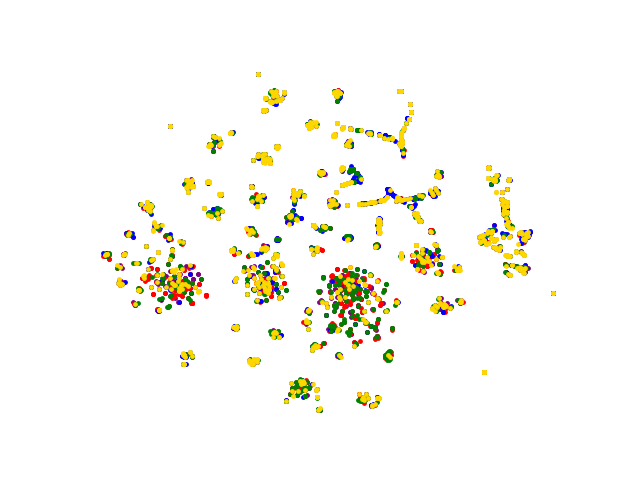}
        \centerline{(e) JODIE}
    \end{minipage}%
        \begin{minipage}[t]{0.125\textwidth}
        \includegraphics[width=1\textwidth]{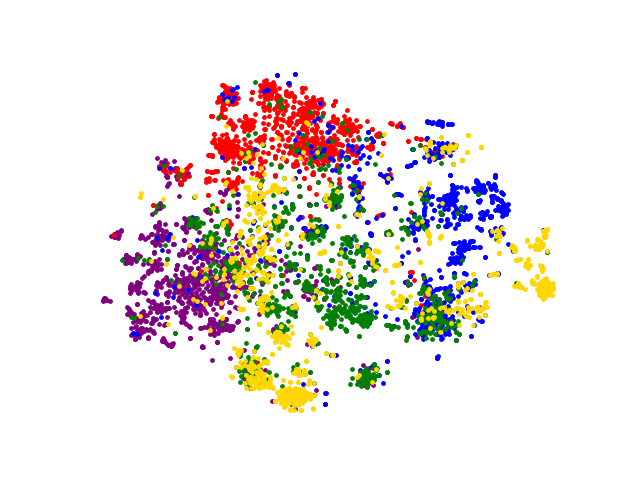}
        \centerline{(f) IJODIE}
    \end{minipage}%
    \begin{minipage}[t]{0.125\textwidth}
        \includegraphics[width=1\textwidth]{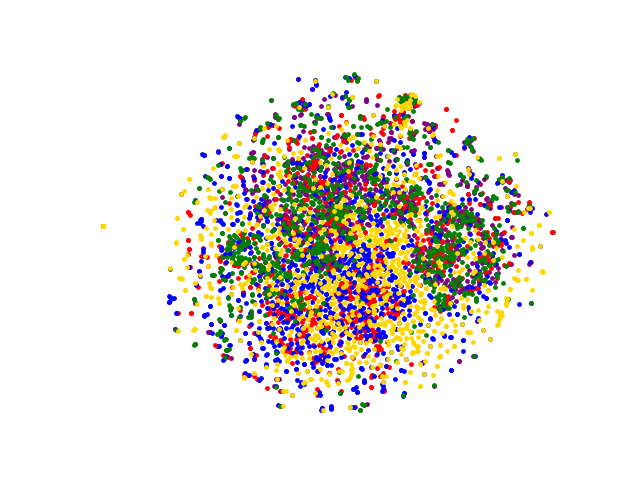}
        \centerline{(g) TREND}
    \end{minipage}%
    \begin{minipage}[t]{0.125\textwidth}
        \includegraphics[width=1\textwidth]{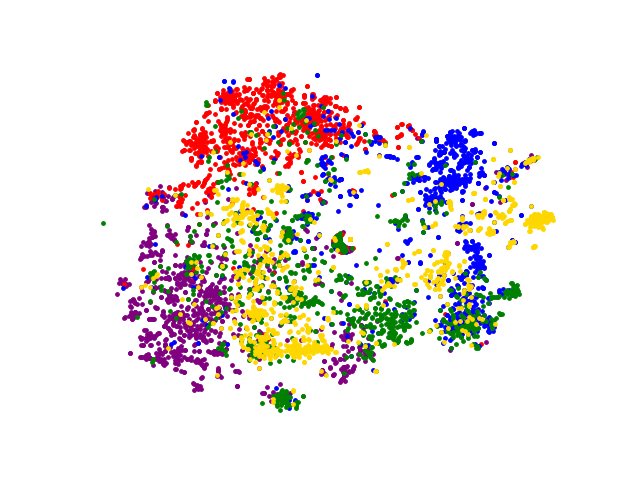}
        \centerline{(h) ITREND}
    \end{minipage}%
    \caption{Visualization of node embeddings using t-SNE algorithm, where the first row shows the results on School dataset ($K=9$) and the second row shows the results on arXivAI dataset ($K=5$).}
    \label{tsne}
\end{figure*}

\begin{figure*}[t]
    \centering
    \begin{minipage}[t]{0.125\textwidth}
        \includegraphics[width=1\textwidth]{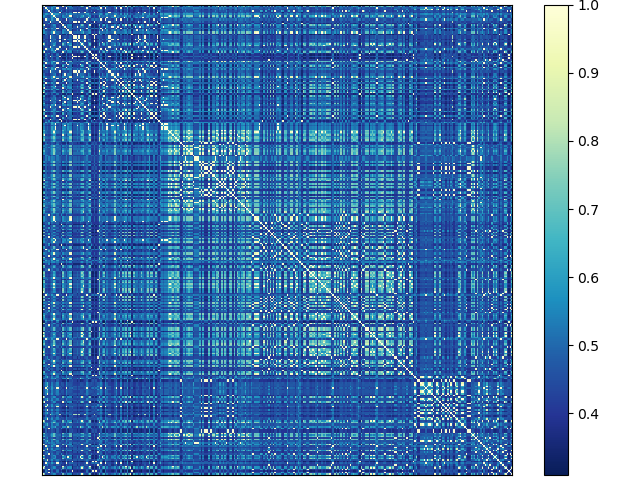}
    \end{minipage}%
    \begin{minipage}[t]{0.125\textwidth}
        \includegraphics[width=1\textwidth]{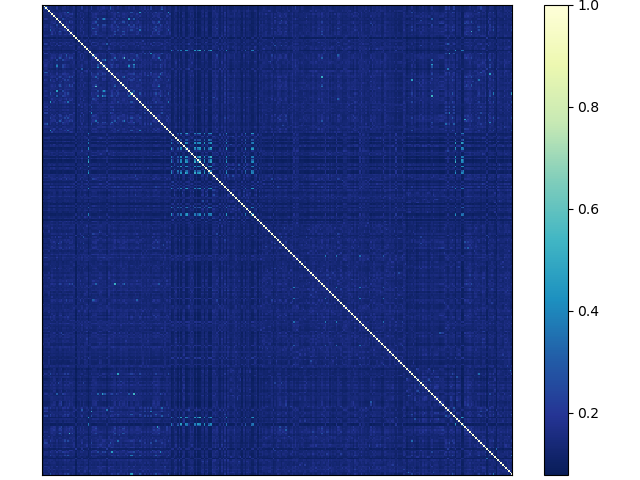}
    \end{minipage}%
    \begin{minipage}[t]{0.125\textwidth}
        \includegraphics[width=1\textwidth]{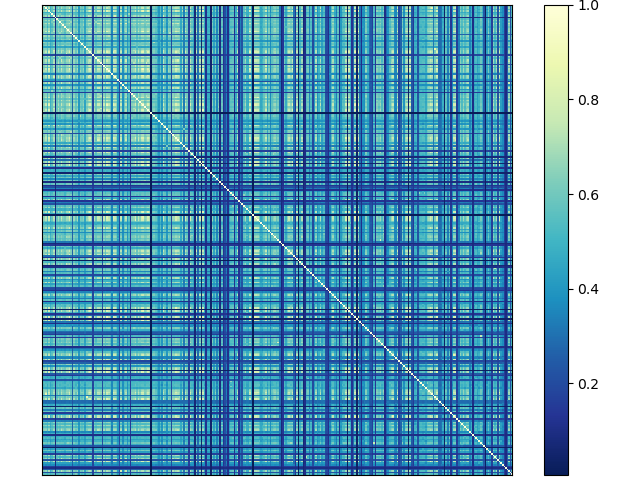}
    \end{minipage}%
    \begin{minipage}[t]{0.125\textwidth}
        \includegraphics[width=1\textwidth]{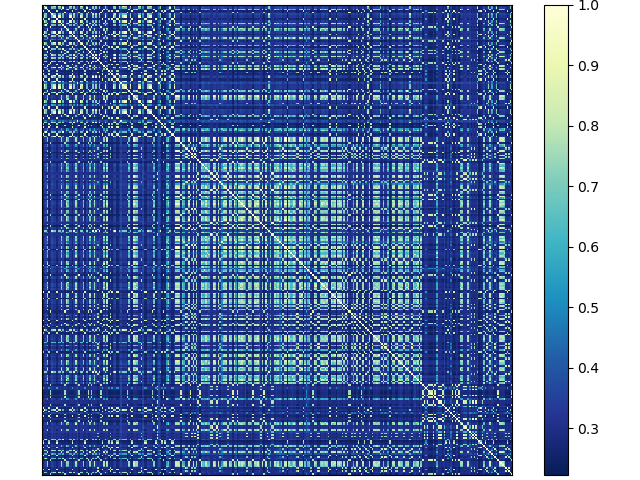}
    \end{minipage}%
    \begin{minipage}[t]{0.125\textwidth}
        \includegraphics[width=1\textwidth]{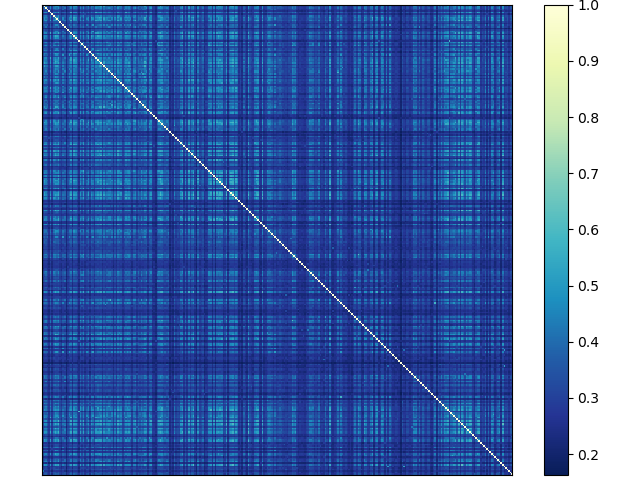}
    \end{minipage}%
        \begin{minipage}[t]{0.125\textwidth}
        \includegraphics[width=1\textwidth]{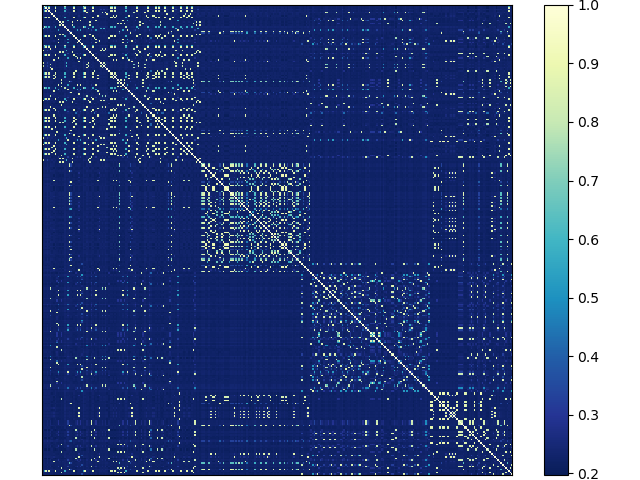}
    \end{minipage}%
    \begin{minipage}[t]{0.125\textwidth}
        \includegraphics[width=1\textwidth]{S2T-P.png}
    \end{minipage}%
    \begin{minipage}[t]{0.125\textwidth}
        \includegraphics[width=1\textwidth]{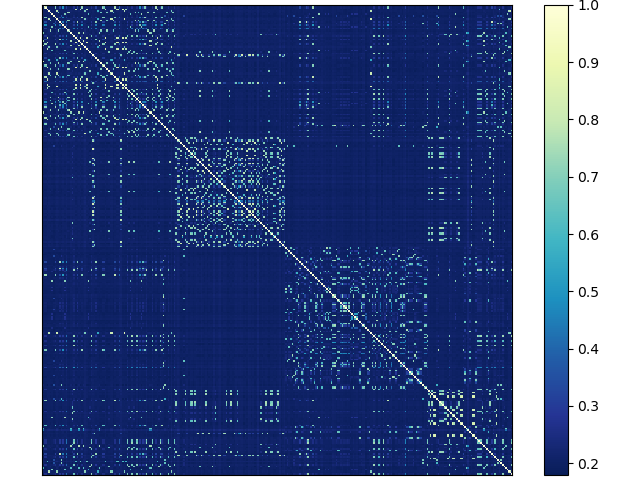}
    \end{minipage}%
    \newline
    \begin{minipage}[t]{0.125\textwidth}
        \includegraphics[width=1\textwidth]{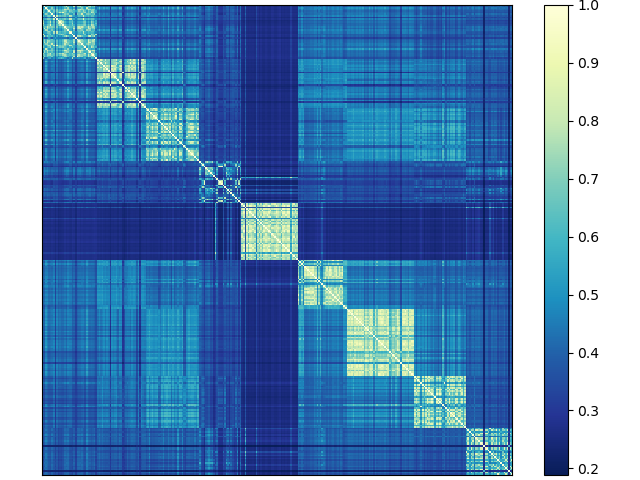}
        \newline
        \centerline{(a) GAE\quad}
    \end{minipage}%
    \begin{minipage}[t]{0.125\textwidth}
        \includegraphics[width=1\textwidth]{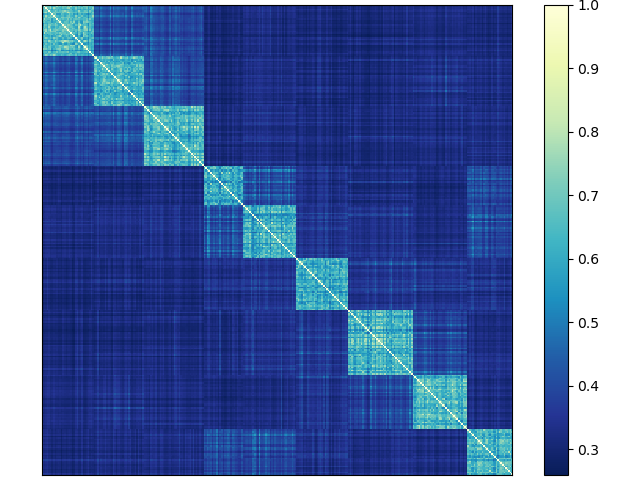}
        \newline
        \centerline{(b) N2V\quad}
    \end{minipage}%
    \begin{minipage}[t]{0.125\textwidth}
        \includegraphics[width=1\textwidth]{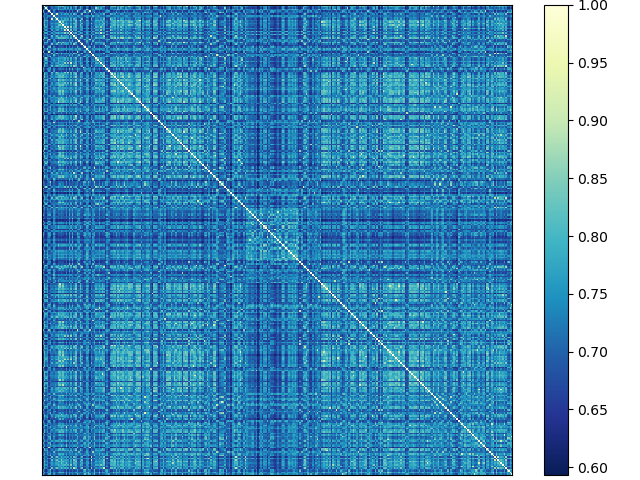}
        \newline
        \centerline{(c) MVGRL\quad}
    \end{minipage}%
    \begin{minipage}[t]{0.125\textwidth}
        \includegraphics[width=1\textwidth]{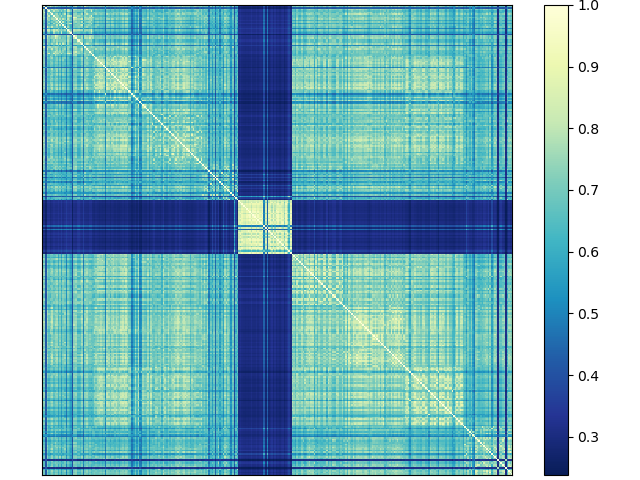}
        \newline
        \centerline{(d) DFCN\quad}
    \end{minipage}%
    \begin{minipage}[t]{0.125\textwidth}
        \includegraphics[width=1\textwidth]{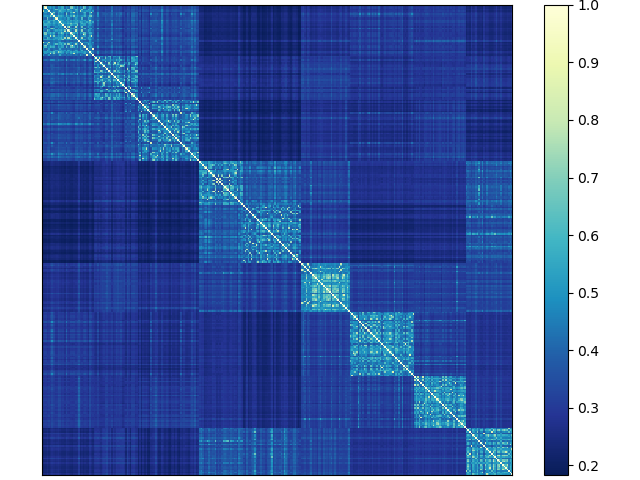}
        \newline
        \centerline{(e) MNCI\quad}
    \end{minipage}%
    \begin{minipage}[t]{0.125\textwidth}
        \includegraphics[width=1\textwidth]{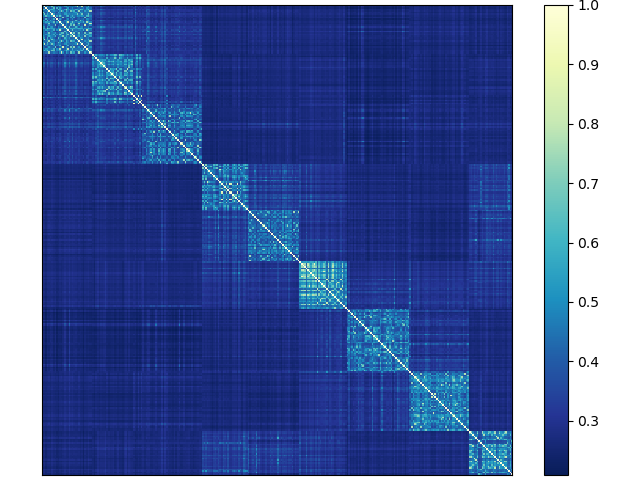}
        \newline
        \centerline{(f) IMNCI\quad}
    \end{minipage}%
    \begin{minipage}[t]{0.125\textwidth}
        \includegraphics[width=1\textwidth]{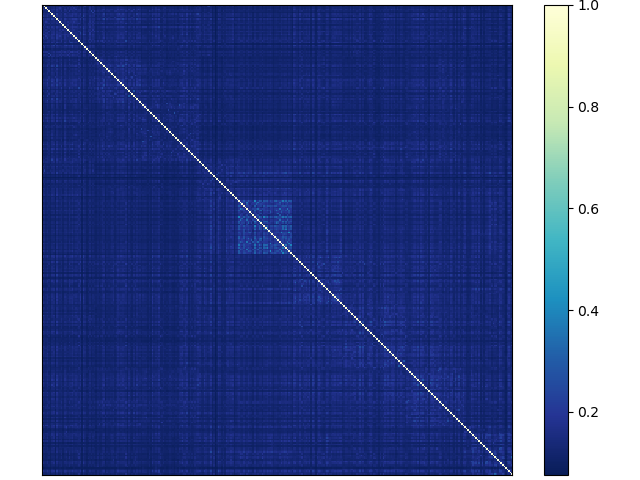}
        \newline
        \centerline{(g) S2T\quad}
    \end{minipage}%
    \begin{minipage}[t]{0.125\textwidth}
    \includegraphics[width=1\textwidth]{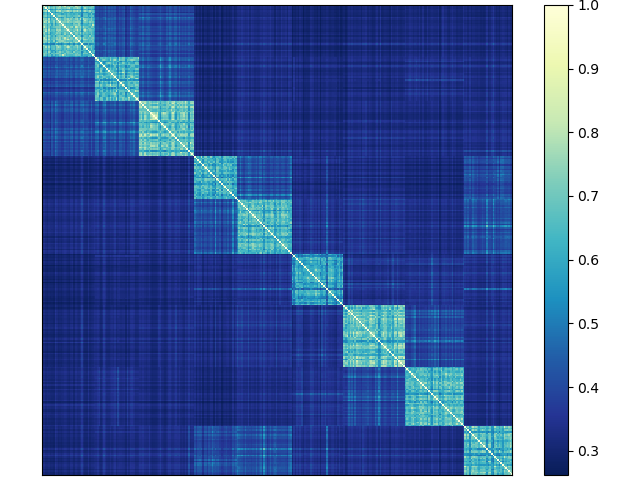}
    \newline
        \centerline{(h) IS2T\quad}
    \end{minipage}%
    \caption{Similarity heat map of node embeddings, where the first row shows the results on Patent dataset ($K$=6) and the second row shows the results on School dataset ($K$=9).}
    \label{sim}
\end{figure*}

As shown in Figure \ref{tsne}, we utilize the t-SNE technology to visualize the final node embeddings. Here we select two dataset, School ($K$ = 9) and arXivAI ($K$ = 5). In the figure, many baseline methods have difficulty in clearly distinguishing node embeddings of different categories, except Deepwalk. In contrast, the node embeddings obtained by our improved methods IJODIE and ITREND have clear boundaries after two-dimensionalization. Especially on the School dataset, since it has only 327 nodes, its clustering index can reach 100\%. In this dataset, each node has a clear affiliation and there are no overlapping clusters. Therefore, for a cluster, the smaller the area it occupies, the clearer the cluster boundary, and the better the clustering performance. 

It is worth mentioning that Deepwalk, although it is modeled on the static graph, is still trained by reading the adjacency list. Deepwalk obtaining node embeddings through random walks can achieve results that exceed those of many graph clustering methods, which is one of the fields worth exploring in the future.

\subsubsection{Similarity visualization}

As shown in Figure \ref{sim}, we use Euclidean distance to calculate the similarity matrix for node embeddings and present it as a heat map. We select two datasets, Patent ($K$=6) and School ($K$=9), and randomly select 300 node embeddings for display. In the figure, nodes belonging to the same cluster are arranged together. If their embeddings are good enough, the similarity between nodes in same cluster should be high, thus forming multiple yellow square areas at the diagonal position. Other positions should be mainly blue, which means that similarities between nodes of this cluster and nodes of other classes are low, that is, the clustering performance is better.

We can find that the heat maps of MNCI and S2T are not clear, but after being improved by our BenchTGC framework, we get clearer similarity distributions. In addition, compared with other methods, our two improved methods are also the clearest in distributions. This fully proves the effectiveness of our proposed BenchTGC framework, which is also consistent with the experimental results above.

\section{Challenges}

\begin{figure}[t]
    \centering
    \includegraphics[width=0.48\textwidth]{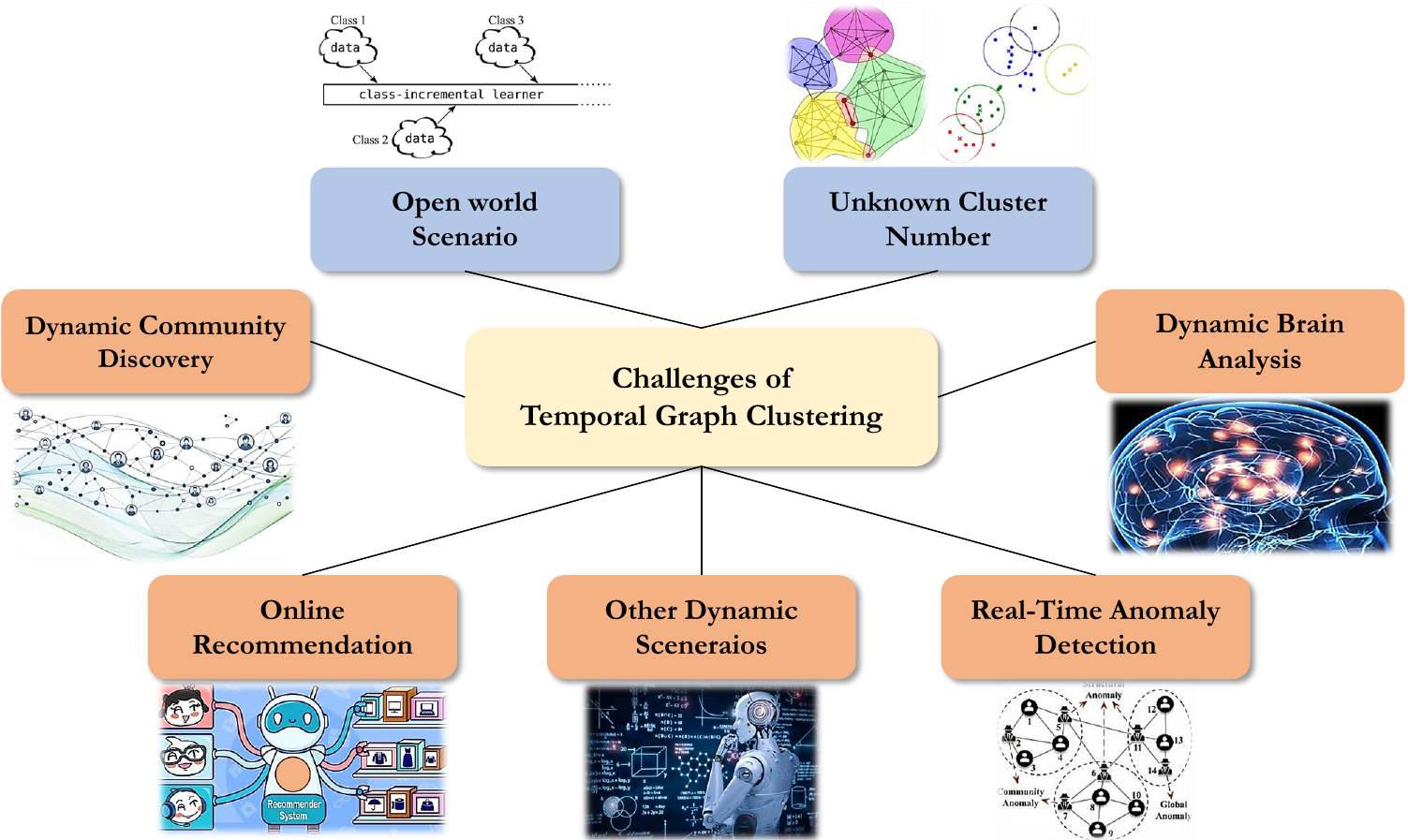}
    \caption{Challenges of TGC, include limitations and applications.}
    \label{challenges}
\end{figure}

Finally, as shown in Figure \ref{challenges}, we talk about the challenges of temporal graph clustering. The limitations and unknown applications that hinder TGC's development are, to some extent, also new opportunities for the future.

\subsection{Limitations}

In addition to the lack of datasets, the development of temporal graph clustering also has other limitations.

\subsubsection{Open World Scenario}

In real world, data flows dynamically. It is difficult to obtain the whole graph, while new data is added one after another. In this case, researchers can hardly train the full graph data and get node embeddings to deploy on the real-time platform. The design of temporal graph clustering methods should consider data input in open scenarios, which is a difficult but attractive challenge.

\subsubsection{Unknown Cluster Number}

In many cases (especially in real applications), we do not know the number of clusters, which means that we cannot specify a specific $K$ value during model training. We should consider how to perform clustering-oriented training without $K$. At the same time, overlapping and non-overlapping clustering situations should also be distinguished.

\subsection{Applications}

Temporal graph clustering can be regarded as graph clustering that is closer to the real world, which means that TGC can match almost all application scenarios of graph clustering well. Additionally, for dynamically changing scenarios \cite{liu2024dynamical}, TGC can better mine the details and make judgments.

\subsubsection{Dynamic Community Discovery}

The most closely related to node clustering is community discovery, which aims to explore potential groups. Researchers utilize community discovery on many real-world scenarios, such as player group mining in game \cite{zhang2023constrained} and epidemic prediction \cite{hy2022temporal}. Such scenarios usually contain rich dynamic information. How to make use of this information will be one of the important issues in the future exploration of TGC.

\subsubsection{Online Recommendation}

The recommendation system is one of the classic applications of graph learning, which focuses on how to recommend products that consumers may buy. Graph clustering can help the recommendation algorithm to further explore or simplify such process \cite{cui2020personalized, liu2024end}, i.e., users or products in the same cluster often have similar characteristics and are the focus of recommendation. How to grasp the temporal information of the ``user-product'' interaction process is a potential application direction.

\subsubsection{Dynamic Brain Analysis}

Among the datasets we use, there is a Brain dataset that records the activation data of neurons in the brain. By analyzing the activation order and frequency of these neurons, clustering can be done well to determine which regions these neurons belong to. This can help researchers analyze changes and evolution in the brain, and temporal graph clustering naturally conforms to the high-frequency dynamic change characteristics of the brain.

\subsubsection{Real-Time Anomaly Detection}

Graph clustering can help detect anomalies by separating normal clusters from abnormal clusters. Similarly, in fields such as social networks and financial transactions, there are a large number of high-frequency interactions, and the temporal information of these interactions will be a very important feature. How to accurately grasp these interactions and provide researchers with real-time detection feedback on anomalies will be one of the potential opportunities.

\subsubsection{Other Dynamic Scenarios}

Currently, many fields begin to use AI technologies to help research or engineering, such as healthcare \cite{si2023multi, wang2024pddgcn} and bioformatics \cite{li2023single, li2025metaq}. Thus there will be more potential scenarios worth exploring. This is also what we want to emphasize: \textbf{The dynamically changing real world are the foundation of temporal graph clustering.} In other words, TGC is not a subdivision of graph clustering, but a more dynamic perspective to observe the real world, providing a new possibility for mining these rich time information.

\section{Conclusion}
Overall, we discuss two major challenges that hinder the development of temporal graph clustering. The BenchTGC Framework and Datasets are proposed for these two challenges, respectively. Combined with experiments, we comprehensively analyze the current problems of static graph clustering and the advantages of temporal graph clustering. Finally, we summarize the limitations and potential applications of temporal graph clustering. We hope BenchTGC can bring new ideas and insights to the graph learning community.

\bibliographystyle{IEEEtran}
\bibliography{tpami}

\vfill

\end{document}